# Sparse Principal Component Analysis for High Dimensional Vector Autoregressive Models

Zhaoran Wang* and Fang Han† and Han Liu‡

**Abstract**

We study sparse principal component analysis for high dimensional vector autoregressive time series under a doubly asymptotic framework, which allows the dimension $d$ to scale with the series length $T$. We treat the transition matrix of time series as a nuisance parameter and directly apply sparse principal component analysis on multivariate time series as if the data are independent. We provide explicit non-asymptotic rates of convergence for leading eigenvector estimation and extend this result to principal subspace estimation. Our analysis illustrates that the spectral norm of the transition matrix plays an essential role in determining the final rates. We also characterize sufficient conditions under which sparse principal component analysis attains the optimal parametric rate. Our theoretical results are backed up by thorough numerical studies.

## 1 Introduction

Principal component analysis is an important technique for dimension reduction and feature extraction and is one of the most employed techniques in multivariate analysis Anderson (1984). Let $X \in \mathbb{R}^d$ be a random vector with covariance matrix $\Sigma$. Principal component analysis aims at recovering the principal subspace corresponding to the top $m$ leading eigenvalues of $\Sigma$. In this paper, we consider principal component analysis for weakly stationary lag $p$ ($p \geq 1$) vector autoregressive time series. For example, let $x_1, \ldots, x_T \in \mathbb{R}^d$ be the realizations of $X_1, \ldots, X_T$ from a lag $p$ vector autoregressive process $(X_t)_{t=-\infty}^{\infty}$ with marginal covariance matrix $\Sigma$. In the introduction section, we take $p = 1$ for simplicity, i.e., $X_t$ satisfies

$$X_{t+1} = AX_t + Z_{t+1} \quad (t \in \mathbb{Z}).$$

---
*Department of Operations Research and Financial Engineering, Princeton University, Princeton, NJ 08544, USA; e-mail: `zhaoran@princeton.edu`.
†Department of Biostatistics, Johns Hopkins University, Baltimore, MD 21205, USA; e-mail: `fhan@jhsph.edu`.
‡Department of Operations Research and Financial Engineering, Princeton University, Princeton, NJ 08544, USA; e-mail: `hanliu@princeton.edu`.

Here $A \in \mathbb{R}^{d \times d}$ is called the transition matrix and the Gaussian noise $Z_{t+1} \sim N_d(0, \Psi)$ is independent from $(X_{t'})_{t'=-\infty}^{t}$. Since the process is weakly stationary, $\Sigma$ satisfies

$$\Sigma = A \Sigma A^{\mathrm{T}} + \Psi.$$

Let $\mathcal{U}$ be the $m$-dimensional ($m = 1, \ldots, d$) principal subspace corresponding to the top $m$ leading eigenvalues of $\Sigma$. We want to find $\widehat{\mathcal{U}}$ to estimate $\mathcal{U}$. In the special case where $A$ is a zero matrix, this problem reduces to the classical principal component analysis for independent Gaussian data.

In this paper we employ a doubly asymptotic framework to analyze multivariate vector autoregressive time series, which allows the time series dimension $d$ to scale with the series length $T$ with $d$ possibly much larger than $T$. In contrast to the classical asymptotic framework which only considers increasing $T$ and fixed $d$, such a doubly asymptotic framework better addresses the challenge of high dimension in enormous real-world applications. For example, in the brain imaging application, an fMRI machine collects $T$ scans of a human brain with $d$ voxels in each scan. Typically, the number of scans $T$ is around hundreds, while the number of voxels $d$ could be tens of thousands. The application of social media stream modeling, e.g., twitter data mining, monitors the tweets from $d$ users across $T$ time units (e.g. minutes). Here $d$ could be millions while $T$ could be hundreds or thousands in a typical setting. In the application of low-frequency stock trading, we have $T$ records of the closing price of stocks, which could be much smaller than the number of stocks $d$.

Such a doubly asymptotic framework poses significant difficulties for theoretical analysis even though it is more realistic. Johnstone and Lu (2009) show that, even in a simplified setting where $A$ is a zero matrix and $\Sigma$ is an identity matrix, the classical principal component analysis can produce inconsistent estimate of the leading eigenvector $u_1$ of $\Sigma$. We need more assumptions to avoid such a curse of dimensionality. One common assumption is that the population leading eigenvector $u_1$ is sparse, i.e., the number of nonzero elements in $u_1$ denoted by $s$ is much smaller than $T$. Under this assumption of sparsity, different methods of sparse principal component analysis have been proposed to handle independent data, including greedy algorithms (d'Aspremont et al., 2008), power methods (Journée et al., 2010; Yuan, 2010; Ma, 2013) and lasso-type methods (Jolliffe et al., 2003; Zou, 2006; Shen and Huang, 2008; Witten et al., 2009; d'Aspremont et al., 2007). Amini and Wainwright (2009); Ma (2013); Paul and Johnstone (2012); Vu and Lei (2012a,b) analyze the theoretical properties of different sparse principal component analysis methods. In detail, we denote $U = (u_1, \ldots, u_m) \in \mathbb{R}^{d \times m}$ to be the matrix consisting of the orthonormal basis of the $m$-dimensional subspace $\mathcal{U}$ corresponding to the top $m$ leading eigenvalues of $\Sigma$. Given that each of $u_1, \ldots, u_m$ has no more than $s$ nonzeros entries, Vu and Lei (2012a,b) show that the sparse principal component analysis estimator achieves the $\{s \log(d)/n\}^{1/2}$ rate of convergence in parameter estimation and prove that this rate is minimax optimal over certain parameter space.

One limitation of these existing sparse principal component analysis theories is that they rely on the assumption that the $T$ realizations $x_1, \ldots, x_T$ are independent. Such an independence assumption does not hold in many real-world applications. In the example of brain imaging, each two adjacent fMRI scans are correlated. Similarly, for stock or twitter data, we can also justify the existence of temporal correlations. Although some results have been established for low dimensional principal component analysis on dependent data (Skinner et al., 1986; Tsay, 2005), there exists no such result in the high dimensional settings. There are some related results on dependent data analysis in high dimensions (Loh and Wainwright, 2012; Fan et al., 2012). However, they mainly focus on high dimensional regression rather than estimating the principal components. In this paper, one of our key proofs adapts from a lemma in Loh and Wainwright (2012).

In this paper, we study sparse principal component analysis for high dimensional stationary time series. More specifically, we treat the transition matrix $A$ as a nuisance parameter and directly apply sparse principal component analysis on the multivariate time series $x_1, \ldots, x_T$ to estimate the principal subspace $\mathcal{U}$ corresponding to the top $m$ leading eigenvalues of $\Sigma$, as if the data are independent. Let $\|A\|_2$ be the spectral norm of the transition matrix $A$. We show that $\|A\|_2$ plays a vital role in determining the final rates of convergence. When $m = 1$, estimating $\mathcal{U}$ is equivalent to estimating the leading eigenvector $u_1$ of $\Sigma$. In our shorter version of this paper (Wang et al., 2013), we provide an explicit rate of convergence for the angle between the leading eigenvector $u_1$ and the estimator $\widehat{u}_1$, i.e., there exists a constant $C > 0$ such that

$$\big|\sin \angle(u_1, \widehat{u}_1)\big| \leq \frac{C}{\lambda_1(\Sigma) - \lambda_2(\Sigma)} \left\{ \frac{s \log(d)}{T} \left( \frac{\|\Sigma\|_2}{1 - \|A\|_2} \right) \right\}^{1/2}$$

with high probability. Here $\lambda_k(\Sigma)$ ($k \geq 1$) is the $k$-th largest eigenvalue of $\Sigma$. In this paper we extend this rate of convergence to the estimation of the $m$-dimensional principal subspace $\mathcal{U}$ with $m \geq 1$. Let $U = (u_1, \ldots, u_m) \in \mathbb{R}^{d \times m}$ be the matrix consisting of the orthonormal basis of $\mathcal{U}$. To estimate $\mathcal{U}$, it suffices to estimate $U$. Let $\widehat{U}$ be the estimator of $U$. We show that, under a certain distance function $D(\cdot, \cdot)$ (detailed definition will be provided later), there exists a constant $C$ such that

$$D(U, \widehat{U}) \leq \frac{mC}{\lambda_m(\Sigma) - \lambda_{m+1}(\Sigma)} \left\{ \frac{s \log(d)}{T} \left( \frac{\|\Sigma\|_2}{1 - \|A\|_2} \right) \right\}^{1/2}$$

with high probability. Moreover, we provide extensions to analyze the lag $p$ ($p \geq 1$) vector autoregressive model.

Our result allows both the eigenvalues of $\Sigma$ and $\|A\|_2$ to scale with $d$. The theoretical results of this paper justifies the popular practices where (sparse) principal component analysis is directly applied on high dimensional time series data, e.g., visualization and feature selection of high dimensional stock data (Feeney and Hester, 1967) and face recognition from videos (Zhao et al., 2003). Other potential applications of this work include the analysis of economic time series (Sims, 1980; Briiggemann and Liitkepohl, 2001; Hatemi-J, 2004), signal processing (de Waele and Broersen, 2003) and brain imaging (Roebroeck et al., 2005).

## 2  Background

In this section, we briefly introduce the vector autoregressive model. Let $A = (A_{jk}) \in \mathbb{R}^{d \times d}$ and $v = (v_1, \ldots, v_d)^{\mathrm{T}} \in \mathbb{R}^d$. For $q > 0$, we define $\|v\|_q = (\sum_{j=1}^{d} |v_j|^q)^{1/q}$. We also define $\|v\|_0$ to be the cardinality of the support of $v$. Let $\lambda_i(A)$ and $\sigma_i(A)$ be the $i$-th largest eigenvalue and singular value of a matrix $A$ respectively. For $q > 0$, we denote $\|A\|_q$ to be the operator norm of $A$. For $q = 1$, $q = 2$ and $q = \infty$, $\|A\|_1 = \max_{1 \leq j \leq d} \sum_{i=1}^{d} |A_{ij}|$, $\|A\|_2 = \sigma_1(A)$ and $\|A\|_\infty = \max_{1 \leq i \leq d} \sum_{j=1}^{d} |A_{ij}|$. The matrix Frobenius norm of $A$ is defined as $\|A\|_F = (\sum_{i=1}^{d} \sum_{j=1}^{d} |A_{ij}|^2)^{1/2}$. The nuclear norm is $\|A\|_* = \sum_{i=1}^{d} \sigma_i(A)$. The $\ell_0$ ball with radius $R$ is $\mathbb{B}_0(R) = \{v \in \mathbb{R}^d : \|v\|_0 \leq R\}$. For two vectors $v_1$ and $v_2$, we define the inner product of $v_1$ and $v_2$ as $\langle v_1, v_2 \rangle = v_1^{\mathrm{T}} v_2$. For matrices $A_1, A_2$, we define the inner product of $A_1$ and $A_2$ as $\langle A_1, A_2 \rangle = \mathrm{tr}(A_1^{\mathrm{T}} A_2)$. For a set $\mathbb{K}$, $|\mathbb{K}|$ is its cardinality. We define $\mathrm{diag}(a_1, \ldots, a_d) \in \mathbb{R}^{d \times d}$ to be the diagonal matrix with $a_1, \ldots, a_d$ being its diagonal entries. We denote the identity matrix in $\mathbb{R}^{d \times d}$ by $I_d$.

An orthonormal matrix is a square matrix with orthonormal unit column vectors, while a semi-orthonormal matrix is a non-square matrix with orthonormal unit column vectors. For a semi-orthonormal matrix $V \in \mathbb{R}^{d \times m}$ with columns $v_1, \ldots, v_m$, we denote $\mathcal{V} = \mathrm{span}(V)$ to be the subspace of $\mathbb{R}^d$ spanned by the columns of $V$. It is easy to see that the projection matrix of $\mathcal{V}$ is $\Pi_{\mathcal{V}} = V V^{\mathrm{T}} \in \mathbb{R}^{d \times d}$. Let $\mathcal{V}^\perp$ be the subspace of $\mathbb{R}^d$ which is orthogonal with $\mathcal{V}$. The projection matrix of $\mathcal{V}^\perp$ is denoted as $\Pi_{\mathcal{V}}^\perp \in \mathbb{R}^{d \times d}$, which satisfies $\Pi_{\mathcal{V}}^\perp = I_d - \Pi_{\mathcal{V}}$.

We use canonical angle to define the distance between $\mathcal{V}_1$ and $\mathcal{V}_2$. Letting $\theta_k(\mathcal{V}_1, \mathcal{V}_2) = \arcsin\{\sigma_k(\Pi_1 \Pi_2^\perp)\}$ ($k = 1, \ldots, m$), we denote $\Theta(\mathcal{V}_1, \mathcal{V}_2) = \mathrm{diag}(\theta_1, \ldots, \theta_m)$ where we abbreviate $\theta_k(\mathcal{V}_1, \mathcal{V}_2)$ as $\theta_k$. For the diagonal matrix $\Theta(\mathcal{V}_1, \mathcal{V}_2)$, we define the $\sin(\cdot) : \mathbb{R}^{d \times d} \to \mathbb{R}^{d \times d}$ operator as $\sin\{\Theta(\mathcal{V}_1, \mathcal{V}_2)\} = \mathrm{diag}\{\sin(\theta_1), \ldots, \sin(\theta_m)\}$. The canonical angle distance between $\mathcal{V}_1, \mathcal{V}_2$ is defined as $\|\sin\{\Theta(\mathcal{V}_1, \mathcal{V}_2)\}\|_F$. More details about the canonical angle can be found in Stewart and Sun (1990) and Bhatia (1997).

For notational simplicity, in the sequel we use $\sin(V_1, V_2)$ to denote $\sin\{\Theta(\mathcal{V}_1, \mathcal{V}_2)\}$, i.e.,

$$\sin(V_1, V_2) = \sin\{\Theta(\mathcal{V}_1, \mathcal{V}_2)\}. \tag{2.1}$$

For the lag 1 vector autoregressive model, we assume $x_1, \ldots, x_T \in \mathbb{R}^d$ are realizations of $X_1, \ldots, X_T$ from a process $(X_t)_{t=-\infty}^{\infty}$ with mean 0 and marginal covariance matrix $\Sigma$:

$$X_{t+1} = A X_t + Z_{t+1} \quad (t \in \mathbb{Z}). \tag{2.2}$$

Here $A$ is the transition matrix. A necessary and sufficient condition for $A$ to be weakly stationary is that $\|A\|_2 < 1$. We assume $(Z_t)_{t=-\infty}^{\infty}$ are independently and identically drawn from $N_d(0, \Psi)$ and $Z_{t+1}$ and $(X_{t'})_{t'=-\infty}^{t}$ are independent. The stationary property implies that

$$\Sigma = A \Sigma A^{\mathrm{T}} + \Psi. \tag{2.3}$$

For $1 \leq j \leq k \leq T$, the covariance between $X_j$ and $X_k$ is

$$\mathrm{Cov}(X_j, X_k) = \underbrace{A \cdots A}_{k-j} \Sigma = A^{k-j} \Sigma, \tag{2.4}$$

where $A^0$ is defined as $I_d$. The lag $p$ vector autoregressive model satisfies

$$X_{t+1} = \sum_{i=1}^{p} A_i X_{t-i+1} + Z_{t+1}, \quad (t \in \mathbb{Z}). \tag{2.5}$$

## 3 Sparse Principal Component Analysis for Multivariate Time Series

Let $x_1, \ldots, x_T$ be the $T$ realizations of $X_1, \ldots, X_T$ from a mean 0 stationary process $(X_t)_{t=-\infty}^{\infty}$. Letting $\Sigma$ be the marginal covariance matrix of $X_t$, we denote $\mathcal{U}$ to be the $m$-dimensional ($m \leq d$) principal subspace corresponding to the $m$ leading eigenvalues of $\Sigma$. We aim to estimate $\mathcal{U}$ based on $x_1, \ldots, x_T$.

We first consider the $m = 1$ case, i.e., estimating the leading eigenvector $u_1$ of $\Sigma$. We assume that $u_1$ satisfies that $\|u_1\|_0 \leq s$. We define the estimator $\widehat{u}_1$ with the following optimization problem:

$$\widehat{u}_1 = \underset{v \in \mathbb{R}^d}{\mathrm{argmax}}\, v^\mathrm{T} S v \quad \text{subject to} \quad v \in \mathbb{S}^{d-1} \cap \mathbb{B}_0(s),$$

where $\mathbb{S}^{d-1} = \{v | v \in \mathbb{R}^d, \|v\|_2 = 1\}$ is the $d$-dimensional Euclidean unit sphere. It is easy to see that, without the sparsity constraint $\mathbb{B}_0(s)$, the solution of (3.1) is the leading eigenvector of $S$.

To extend this estimator in (3.1) to estimate the $m$-dimensional principal subspace ($m = 1, \ldots, d$), we adapt the estimator proposed by Vu and Lei (2012b). More specifically, we define the $\ell_0$ pseudo-norm of $U = (u_1, \ldots, u_m) \in \mathbb{R}^{d \times m}$ as

$$\|U\|_0 = \sum_{j=1}^{d} \mathbb{I}(U_{j*} \neq 0), \tag{3.1}$$

where $U_{j*}$ is the $j$-th row of $U$. The indicator function $\mathbb{I}(U_{j*} \neq 0) = 1$ if and only if $U_{j*}$ is not a zero vector. $\|U\|_0$ can also be viewed as the $\ell_0$ pseudo-norm of the column vector with entries $\|U_{j*}\|_2$ ($j = 1, \ldots, d$). When $m = 1$, $\|U\|_0$ is the number of nonzero elements of $u_1$. This pseudo-norm is coordinate-independent, i.e., $\|U\|_0 = \|UO\|_0$ for any orthonormal matrix $O \in \mathbb{R}^{m \times m}$. Let $\mathbb{V}$ is the set of $d \times m$ semi-orthonormal matrices and $\mathrm{span}(V)$ is the subspace spanned by the columns of $V$. We then define the corresponding collection of $m$-dimensional principal subspaces with sparsity constraints to be

$$\mathcal{P}_0(s, m) = \{\mathrm{span}(V) : V \in \mathbb{V} \subset \mathbb{R}^{d \times m},\ \|V\|_0 \leq s\}. \tag{3.2}$$

We estimate the principal subspace $\mathcal{U}$ using the estimator

$$\widehat{\mathcal{U}} = \underset{\mathcal{V} \in \mathcal{P}_0(s,m)}{\operatorname{argmin}} \frac{1}{T} \sum_{i=1}^{T} \|(I_d - \Pi_\mathcal{V}) X_i\|_2^2,$$

where $\Pi_\mathcal{V}$ is the projection matrix of the subspace $\mathcal{V}$. As has been shown in Vu and Lei (2012b), the above estimator is equivalent to the following optimization problem

$$\widehat{U} = \operatorname{argmax} \langle S, VV^\mathrm{T} \rangle \quad \text{subject to} \quad V \in \mathbb{V}, \ \|V\|_0 \leq s. \tag{3.3}$$

It is easy to see that (3.1) is a special case of (3.3) when $m = 1$. As is shown in Vu and Lei (2012b), even though the estimators in (3.1) and (3.3) are computationally intractable since they are essentially combinatoric, theoretical results obtained for these estimators provide deeper insights of the sparse principle component analysis problem and can be used to calibrate other more practical algorithms. In §6, we also propose a two-step heuristic algorithm to approximately compute the the estimators in (3.1) and (3.3), which works well in our numerical simulations.

## 4 Theoretical Properties

We provide theoretical properties of the sparse principal component analysis estimators in (3.1) and (3.3) for vector autoregressive time series. In particular, we provide the explicit nonasymptotic upper bound for the rates of convergence in parameter estimation. All the proofs are provided in the appendix.

To describe our results, we define the model class $\mathcal{M}(s, m, \Sigma)$ as follows:

$$\mathcal{M}(s, m, \Sigma) = \{X : X \sim N_d(0, \Sigma), \ \mathcal{U} \in \mathcal{P}_0(s, m), \lambda_m(\Sigma) > \lambda_{m+1}(\Sigma)\}.$$

Here we remind that $\mathcal{U}$ is the $m$-dimensional principal subspace corresponding to the top $m$ leading eigenvalues of $\Sigma$ and $\mathcal{P}_0(s, m)$ is the collection of subspaces with sparsity constraints as defined in (3.2).

Our first result states that for vector autoregressive time series with lag 1, the leading eigenvector estimator $\widehat{u}_1$ obtained from (3.1) approximates $u_1$ in an explicit rate of convergence.

**Theorem 4.1.** Provided that $X_1, \ldots, X_T \in \mathcal{M}(s, 1, \Sigma)$ are from the lag 1 vector autoregressive process $(X_t)_{t=-\infty}^{\infty}$ described in (2.2), the estimator $\widehat{u}_1$ in (3.1) has the following rate of convergence:

$$\left| \sin \angle (u_1, \widehat{u}_1) \right| = O_P \left[ \frac{1}{\lambda_1(\Sigma) - \lambda_2(\Sigma)} \left\{ \frac{s \log(d)}{T} \left( \frac{\|\Sigma\|_2}{1 - \|A\|_2} \right) \right\}^{1/2} \right]. \tag{4.1}$$

Moreover, we extend this result to estimating the principal subspace $\mathcal{U}$ corresponding to the top $m$ ($m \geq 1$) leading eigenvalues of $\Sigma$. We show that $\widehat{U}$ obtained from (3.3) is a consistent estimator even when $d$ is almost exponentially larger than $T$.

**Theorem 4.2.** Assuming that $X_1, \ldots, X_T \in \mathcal{M}(s, m, \Sigma)$ are from the lag 1 vector autoregressive process $(X_t)_{t=-\infty}^{\infty}$ described in (2.2), the estimator $\widehat{U}$ in (3.3) satisfies

$$\left\| \sin\left(U, \widehat{U}\right) \right\|_F = O_P \left[ \frac{m}{\lambda_m(\Sigma) - \lambda_{m+1}(\Sigma)} \left\{ \frac{s \log(d)}{T} \left( \frac{\|\Sigma\|_2}{1 - \|A\|_2} \right) \right\}^{1/2} \right], \tag{4.2}$$

where $\sin\left(U, \widehat{U}\right)$ is defined in (2.1).

**Remark 4.3.** The rates of convergence in (4.1) and (4.2) depend on $\Sigma$ and $A$, where $A$ characterizes the degree of temporal dependence. The result in (4.1) shows that, when $\|A\|_2$ does not scale with $(T, d, s)$, $\widehat{u}_1$ attains the optimal parametric rate of convergence for estimating the leading eigenvector $u_1$ over certain parameter space (More details about the minimax optimality can be found in Vu and Lei (2012a) ). Moreover, when $\|A\|_2$ does not scale with $(T, d, s)$ and $m$ is fixed, $\widehat{U}$ attains the optimal rate of subspace estimation (Vu and Lei, 2012b).

In addition, the above results can be extended to the lag $p$ vector autoregressive model as in the following corollary.

**Corollary 4.4.** Given that $X_1, \ldots, X_T \in \mathcal{M}(s, m, \Sigma)$ are from the lag $p$ vector autoregressive process $(X_t)_{t=-\infty}^{\infty}$ described in (2.5), the estimator $\widehat{U}$ in (3.3) satisfies

$$\left\| \sin\left(U, \widehat{U}\right) \right\|_F = O_P \left[ \frac{m}{\lambda_m(\Sigma) - \lambda_{m+1}(\Sigma)} \left\{ \frac{s \log(pd)}{T} \left( \frac{\|\widetilde{\Sigma}\|_2}{1 - \|\widetilde{A}\|_2} \right) \right\}^{1/2} \right], \tag{4.3}$$

where $\Sigma$ is the covariance matrix of $X_t$, $\widetilde{\Sigma}, \widetilde{A} \in \mathbb{R}^{pd \times pd}$ are defined as

$$\widetilde{\Sigma} = \begin{pmatrix} \Sigma & \mathrm{Cov}(X_1, X_2) & \cdots & \mathrm{Cov}(X_1, X_p) \\ \mathrm{Cov}(X_2, X_1) & \Sigma & \cdots & \vdots \\ \vdots & \ddots & \ddots & \vdots \\ \mathrm{Cov}(X_p, X_1) & \cdots & \cdots & \Sigma \end{pmatrix}, \quad \widetilde{A} = \begin{pmatrix} A_1 & A_2 & \cdots & A_{p-1} & A_p \\ I_d & 0 & \cdots & 0 & 0 \\ 0 & I_d & \cdots & 0 & 0 \\ \vdots & \ddots & \ddots & \ddots & \vdots \\ 0 & 0 & \cdots & I_d & 0 \end{pmatrix}.$$

## 5 Implications of the Main Results

In this section, we use several concrete examples to gain more insights of Theorem 4.1 and Theorem 4.2. We also characterize sufficient conditions under which sparse principal component analysis on vector autoregressive time series achieves the same parametric rate as for the independent data. For simplicity, we focus on the lag 1 vector autoregressive model, while the corresponding results for the lag $p$ vector autoregressive model follow directly due to Corollary 4.4.

We consider the estimator of the $m$-dimensional principal subspace in (3.3), where $X_1, \ldots, X_T \in \mathcal{M}(s, m, \Sigma)$ come from the lag 1 vector autoregressive process $(X_t)_{t=-\infty}^{\infty}$ in (2.5). The obtained rate of convergence in (4.2) can be represented as

$$\|\sin(U,\widehat{U})\|_F = O_P\left[\kappa\left\{\frac{s\log(d)}{T}\right\}^{1/2}\right],$$

where

$$\kappa = \frac{m}{\lambda_m(\Sigma) - \lambda_{m+1}(\Sigma)}\left(\frac{\|\Sigma\|_2}{1-\|A\|_2}\right)^{1/2}. \tag{5.1}$$

The stationary property of the lag 1 vector autoregressive model reads as

$$\Sigma = A\Sigma A^{\mathrm{T}} + \Psi. \tag{5.2}$$

Bitmead (1981) shows that (5.2) is a special case of the discrete-time Lyapunov equation with an explicit solution

$$\Sigma = \sum_{i=0}^{\infty}(A)^i \Psi (A^{\mathrm{T}})^i. \tag{5.3}$$

Therefore, we are able to give the simplified forms of $\kappa$ in the following examples.

**Example 5.1.** Assuming $A = \rho I_d$ with $\rho < 1$, we know $\|A\|_2 = \rho$ and $\Sigma = \Psi/(1-\rho^2)$. Since $\lambda_m(\Sigma) = \lambda_m(\Psi)/(1-\rho^2)$ and $\lambda_{m+1}(\Sigma) = \lambda_{m+1}(\Psi)/(1-\rho^2)$, $\kappa$ can be written as

$$\kappa = \frac{m\{\lambda_1(\Psi)\}^{1/2}}{\lambda_m(\Psi) - \lambda_{m+1}(\Psi)}. \tag{5.4}$$

**Example 5.2.** When $A$ is symmetric and $\Psi = \tau I_d$ with $\tau \in \mathbb{R}$, using singular value decomposition we have $A = Q\Lambda Q^{-1}$. Here $Q \in \mathbb{R}^{d\times d}$, $QQ^{\mathrm{T}} = I_d$ and $\Lambda$ is the diagonal matrix with $\Lambda_{ii} = \lambda_i(A)$. We then have

$$\Sigma = \tau Q \sum_{i=0}^{\infty} \Lambda^i Q^{-1}.$$

Thus, we get $\lambda_m(\Sigma) = \tau/\{1-\lambda_m(A)\}$, $\lambda_{m+1}(\Sigma) = \tau/\{1-\lambda_{m+1}(A)\}$. We have

$$\kappa = \frac{\tau^{-1/2}m\{1-\lambda_m(A)\}\{1-\lambda_{m+1}(A)\}}{\{\lambda_m(\widetilde{A}) - \lambda_{m+1}(A)\}\left[\{1-\lambda_1(A)\}\{1-|\lambda_1(A)|\}\right]^{1/2}}. \tag{5.5}$$

For the specific examples in (5.4)–(5.5) and the general case in (5.1), we have the following results: Assuming that $X_1,\ldots,X_T \in \mathcal{M}(s,m,\Sigma)$ are from the lag 1 vector autoregressive process in (2.5), if $\kappa = O(1)$ when both $d$ and $T$ increase (with $m$ fixed), the subspace estimator in (3.3) achieves the same optimal parametric rate as in the independent data case. Moreover, as long as $\kappa = o\big[\{T/\log(d)\}^{1/2}\big]$e, the subspace estimator in (3.3) is still consistent.

# 6 Experiments

We provide experimental results to back up the theoretical analysis. As mentioned in §3, solving (3.1) and (3.3) is computationally intractable due to their combinatoric nature. To approximately solve (3.1) and (3.3), we utilize a two-step heuristics. In the first step, we obtain an approximate solution using the truncated power method (Yuan and Zhang, 2011) and iterative deflation method (Mackey, 2009). In the second step, we conduct a local exhaustive search to refine the solution. More details are provided below.

To approximately calculate $\widehat{u}_1$ in (3.1), we use the truncated power method in the first step. The truncated power method is an iterative procedure. During each iteration, it uses the power method to estimate the leading eigenvector and truncate the solution to be sparse. More details can be found in Yuan and Zhang (2011). We denote the output of this step as $\widetilde{u}_1$. In the second step, we solve

$$\widehat{u}_1 = \operatorname*{argmax}_{v \in \mathbb{R}^d} v^\mathrm{T} S v \quad \text{subject to} \quad v \in \mathbb{S}^{d-1}, \ \operatorname{supp}(v) \subseteq \operatorname{supp}(\widetilde{u}_1),$$

which amounts to singular value decomposition of the submatrix of $S$, which corresponds to each subset of $\operatorname{supp}(\widetilde{u}_1)$. On some smaller scale problems, we compare this two-stage heuristic algorithm with the exhaustive search solution and found that they are in general the same.

To approximately calculate $\widehat{U}$ in (3.3), we use a similar strategy. In the first step we use the iterative deflation method along with the truncated power method to estimate the principle subspace corresponding to the top $m$ leading eigenvectors. More details about the iterative deflation method can be found in(Mackey, 2009). Let the output of the first step be $\widetilde{U} = (\widetilde{u}_1, \ldots, \widetilde{u}_m)$. In the second step, we solve

$$\widehat{U} = \operatorname{argmax} \left\langle S, VV^\mathrm{T} \right\rangle \quad \text{subject to} \quad V \in \mathbb{V}, \ \operatorname{supp}(V) \subseteq \operatorname{supp}(\widetilde{u}_1) \cup \cdots \cup \operatorname{supp}(\widetilde{u}_m).$$

Here $\operatorname{supp}(V)$ is defined as $\{j : V_{j*} \neq 0\}$ where $V_{j*}$ is the $j$-th row of $V$. Again, this is equivalent to solving a singular value decomposition of the submatrix of $S$ indexed by each subset of $\operatorname{supp}(\widetilde{u}_1) \cup \cdots \cup \operatorname{supp}(\widetilde{u}_m)$.

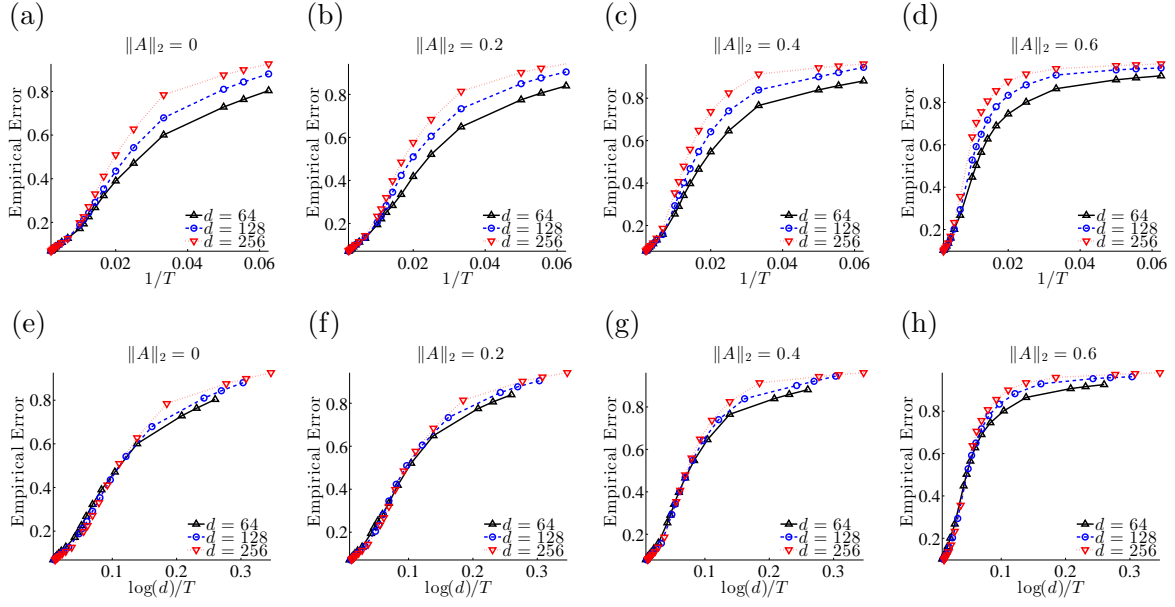

Figure 1: Empirical error of estimating $u_1$ with fixed $\|A\|_2 = 0, 0.2, 0.4, 0.6$ and $d = 64, 128, 256$. (A) The empirical error plotted against $1/T$. (B) The empirical error plotted against $\log(d)/T$.

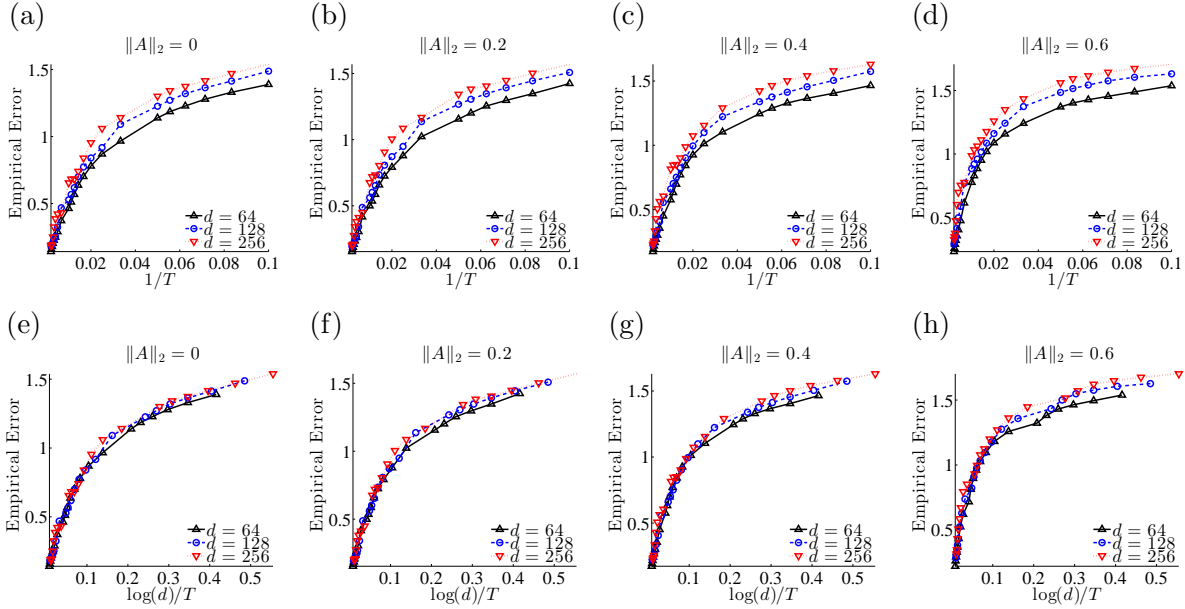

Figure 2: Empirical error of subspace estimation ($m = 4$) with fixed $\|A\|_2 = 0, 0.2, 0.4, 0.6$ and $d = 64, 128, 256$. (A) The empirical error plotted against $1/T$. (B) The empirical error plotted against $\log(d)/T$.

We verify our theoretical results on synthetic datasets. For estimating the leading eigenvector,

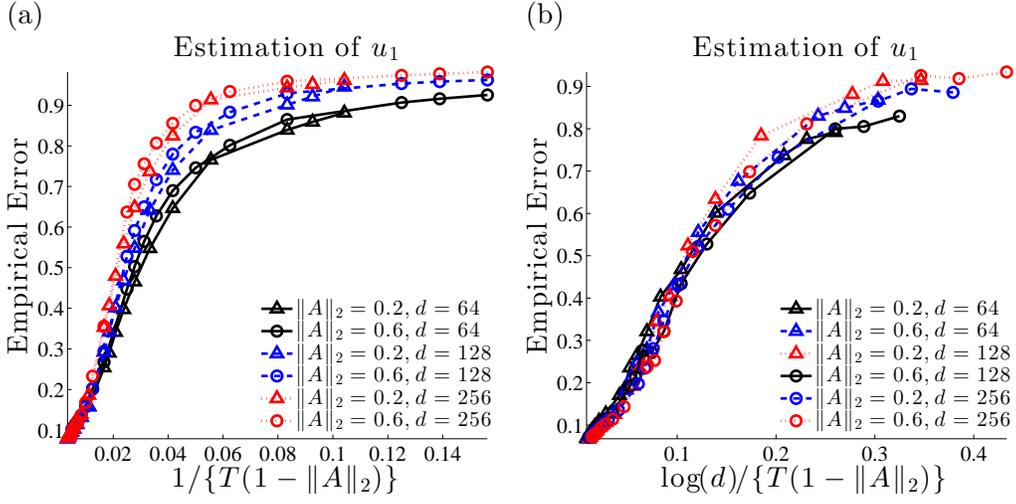

Figure 3: Empirical error for estimating $u_1$ with $\|A\|_2 = 0, 0.2, 0.4, 0.6$ and $d = 64, 128, 256$. (A) The empirical error plotted against $1/\{T(1-\|A\|_2)\}$. (B) The empirical error plotted against $\log(d)/\{T(1-\|A\|_2)\}$.

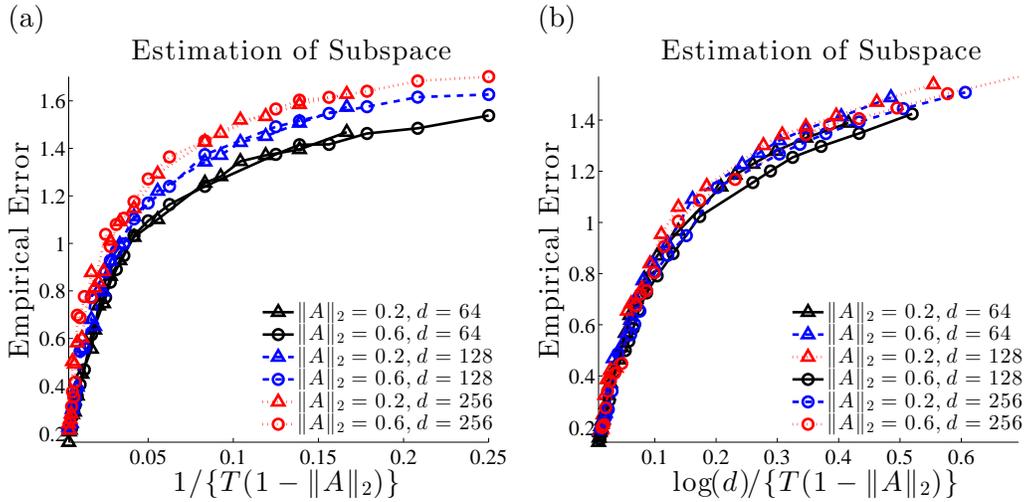

Figure 4: Empirical error for subspace estimation ($m = 4$) with $\|A\|_2 = 0, 0.2, 0.4, 0.6$ and $d = 64, 128, 256$. (A) The empirical error plotted against $1/\{T(1-\|A\|_2)\}$. (B) The empirical error plotted against $\log(d)/\{T(1-\|A\|_2)\}$.

we set $s = 10$, $d = 64, 128, 256$ and $T = 16, 18, 20, 30, \ldots, 100, 150, \ldots, 500$. We create the transition matrix $A$ with prefixed largest singular values $\sigma_1(A) = 0, 0.2, 0.4, 0.6$. To generate the marginal covariance $\Sigma$ of $(X_t)_{t=-\infty}^{\infty}$, we assume the two largest eigenvalues are $\lambda_1(\Sigma) = 10$ and $\lambda_2(\Sigma) = 5$. The rest 48 eigenvalues are all equal to 1. The top two leading eigenvectors $u_1, u_2$ of $\Sigma$ are set to have nonzero entries in their first and second 10 dimensions. The nonzero entries of $u_1$ and $u_2$ are all set to be $10^{-1/2}$. We then generate $\Sigma$ as $\Sigma = \sum_{i=1}^{2} \{\lambda_i(\Sigma) - 1\} u_i u_i^{\mathrm{T}} + I_d$.

Similarly, for subspace estimation, we use the same settings of $d$, $s$ and $T$. To generate $\Sigma$, we assume the 5 largest eigenvalues are $\lambda_1(\Sigma) = 100$, $\lambda_2(\Sigma) = 50$, $\lambda_3(\Sigma) = 30$, $\lambda_4(\Sigma) = 15$ and $\lambda_5(\Sigma) = 2$. The rest 45 eigenvalues are all set to be 1. We generate orthonormal $u_1, \ldots, u_4$ with nonzero entries in its first 10 dimensions and $u_5$ with all-zero entries in the first 10 dimensions. We generate $\Sigma$ from $\Sigma = \sum_{i=1}^{5} \{\lambda_i(\Sigma) - 1\} u_i u_i^{\mathrm{T}} + I_d$.

Using $\Sigma$ and $A$, by the stationary property in (2.3), we can obtain the noise matrix $\Psi = \Sigma - A\Sigma A^{\mathrm{T}}$. The realizations are then generated by (2.2). We repeat the experiments for $1,000$ times and report the averaged results in Appendix 2.

We use $|\sin(u_i, \widehat{u}_i)|$ to evaluate the leading eigenvector estimator in (3.1). In Figure 1, we plot the empirical error against the relevant parts in theoretical upper bound: $1/T$ (in (a)–(d)) and $s\log(d)/T$ (in (e)–(h)) for different settings of $\|A\|_2$ and $d$. From Figure 1 we see that for fixed $\|A\|_2$ and $d$, the empirical error increases with both $1/T$ and $\log(d)/T$, which is consistent with the theoretical result in (4.1). For subspace estimation, we use $\|\sin(U, \widehat{U})\|_F$ as the measure of empirical error for the subspace estimator in (3.3) and observe similar results, i.e., the empirical error increases with both $1/T$ (in (a)–(d)) and $s\log(d)/T$ (in (e)–(h)) with fixed $\|A\|_2$ and $d$. The only difference between (a)–(d) and (e)–(h) is that the horizontal axes in (a)–(d) are rescaled by $\log(d)$ in (e)–(h). We observe that the curves of empirical error plotted against the rescaled axes all stack up, which indicates that the same $\log(d)/T$ leads to a similar empirical error. Hence this observation suggests the $\log(d)/T$ part in our obtained rate of convergence is sharp.

In Figure 3 and Figure 4, we plot the empirical error of the leading eigenvector estimator in (3.1) and subspace estimator in (3.3) against $1/\{T(1-\|A\|_2)\}$ (in (a)) and $\log(d)/\{T(1-\|A\|_2)\}$ (in (b)) respectively. In (a) of Figure 3 and Figure 4, the curves with the same $d$ (of the same color) stack up. This suggests that for fixed $d$ and arbitrary $T$ and $\|A\|_2$, the same $1/\{T(1-\|A\|_2)\}$ indicates similar empirical error. Therefore the $1/\{T(1-\|A\|_2)\}$ part in our obtained rate of convergence is sharp. Parts (b) of both Figure 3 and Figure 4 rescale the horizontal axes of (a) by $\log(d)$. We observe that the curves of the same $d$ in (a) further overlap with the other curves with different $d$ in (b). This observation indicates that the same $\log(d)/\{T(1-\|A\|_2)\}$ leads to similar empirical error, which double confirms that the $\log(d)/\{T(1-\|A\|_2)\}$ part in our obtained rate is sharp.

# 7 Discussion

We study sparse principal component analysis for high dimensional vector autoregressive models. Under a doubly asymptotic framework, we provide explicit rates of convergence for the estimation of principal subspace. Our theoretical results characterize the impact of the temporal dependence on the estimation accuracy and quantify it by the spectral norm of the transition matrix. We also provide sufficient conditions, under which our estimator attains the optimal parametric rate. Although there have been results on sparse principal component analysis for independent data, to the best of our knowledge, this is the first work analyzing the theoretical performance of sparse principal component analysis for high dimensional multivariate time series. Our results provide theoretical justification for applications where sparse principal component analysis is directly applied on high dimensional time series data, e.g., finance, genomics and brain imaging. Besides these theoretical results, our two-stage heuristic procedure to approximate the computational intractable estimators is of its independent interest.

# Appendix A

### Proof of Theorem 4.1

We first prove the rate of convergence in (4.1) for estimating the leading eigenvector $u_1$. In the following, we present several lemmas. The first lemma provides some useful properties of $\sin \angle(u_1, \widehat{u}_1)$, which quantifies the distance between $u_1$ and $\widehat{u}_1$. In the sequel, we assume that the assumptions in Theorem 4.1 hold.

**Lemma .1.** For $u_1$ and $\widehat{u}_1$, we have

$$\sin \angle(u_1, \widehat{u}_1) \leq \frac{2}{\lambda_1 - \lambda_2} \sup_{v \in \mathbb{S}^{d-1} \cap \mathbb{B}_0(2s)} |v^{\mathrm{T}}(\Sigma - S)v|. \tag{.1}$$

*Proof.* Let $\lambda_1 \geq \cdots \geq \lambda_d$ be the eigenvalues of $\Sigma$, with $u_1, \ldots, u_d$ being the basis of the corresponding principal subspace. We have $\Sigma = \lambda_1 u_1 u_1^{\mathrm{T}} + \Phi_0$. We have

$$\begin{aligned}\Phi_0 &= \Sigma - \lambda_1 u_1 u_1^{\mathrm{T}} = \Sigma - \lambda_1 u_1 u_1^{\mathrm{T}} - \lambda_1 u_1 u_1^{\mathrm{T}} + \lambda_1 u_1 u_1^{\mathrm{T}} \\ &= \Sigma - u_1 u_1^{\mathrm{T}} \Sigma - \Sigma u_1 u_1^{\mathrm{T}} + u_1 (u_1^{\mathrm{T}} \Sigma u_1) u_1^{\mathrm{T}} = (I_d - u_1 u_1^{\mathrm{T}}) \Sigma (I_d - u_1 u_1^{\mathrm{T}}).\end{aligned}$$

For any $u \in \mathbb{S}^{d-1}$, we have

$$\begin{aligned}\langle \Sigma, u_1 u_1^{\mathrm{T}} - uu^{\mathrm{T}} \rangle &= \langle \Sigma, u_1 u_1^{\mathrm{T}} \rangle - \langle \lambda_1 u_1 u_1^{\mathrm{T}} + \Phi_0, uu^{\mathrm{T}} \rangle \\ &= \lambda_1 - \lambda_1 \langle u_1, u \rangle^2 - \langle \Phi_0, uu^{\mathrm{T}} \rangle = \lambda_1 - \lambda_1 \langle u_1, u \rangle^2 - u^{\mathrm{T}}(I_d - u_1 u_1^{\mathrm{T}}) \Sigma (I_d - u_1 u_1^{\mathrm{T}}) u.\end{aligned}$$

Now we consider the unit vector $a = (I_d - u_1 u_1^T) u / \|(I_d - u_1 u_1^T) u\|_2 \in \mathbb{R}^d$, which is orthogonal with $u_1$. We have $a = \sum_{j=2}^d a_j u_j$, where $a_j = a^T u_j$ $(j = 2, \ldots, d)$. Therefore, we obtain

$$a^T \Sigma a = a^T \left( \lambda_1 u_1 u_1^T + \sum_{j=2}^d \lambda_j u_j u_j^T \right) a = \sum_{j=2}^d \lambda_j a_j^2 \leq \lambda_2,$$

which indicates

$$u^T (I_d - u_1 u_1^T) \Sigma (I_d - u_1 u_1^T) u \leq \lambda_2 \|(I_d - u_1 u_1^T) u\|_2^2 = \lambda_2 u^T (I_d - u_1 u_1^T) u = \lambda_2 \bigl(1 - \langle u_1, u \rangle^2\bigr).$$

Therefore we get $\langle \Sigma, u_1 u_1^T - u u^T \rangle \geq (\lambda_1 - \lambda_2)\bigl(1 - \langle u_1, u \rangle^2\bigr)$. Let $u = \widehat{u}_1$, we have

$$\langle \Sigma, u_1 u_1^T - \widehat{u}_1 \widehat{u}_1^T \rangle \geq (\lambda_1 - \lambda_2) \sin^2 \angle (u_1, \widehat{u}_1).$$

Moreover, since $u_1$ is feasible and $\widehat{u}_1$ is defined as $\widehat{u}_1 = \operatorname{argmax}_{v \in \mathbb{S}^{d-1} \cap \mathbb{B}_0(s)} v^T S v$, we have $\langle S, u_1 u_1^T - \widehat{u}_1 \widehat{u}_1^T \rangle \leq 0$. Thus, we arrive at

$$\sin^2 \angle (u_1, \widehat{u}_1) \leq \frac{1}{\lambda_1 - \lambda_2} \langle \Sigma, u_1 u_1^T - \widehat{u}_1 \widehat{u}_1^T \rangle \leq \frac{1}{\lambda_1 - \lambda_2} \langle \Sigma - S, u_1 u_1^T - \widehat{u}_1 \widehat{u}_1^T \rangle.$$

Let $P$ be the diagonal matrix with diagonal values 1 if and only if the corresponding entries in $u_1$ or $\widehat{u}_1$ are zero. Therefore, there are at most $2s$ nonzero values in $P$. Then we obtain

$$\sin^2 \angle (u_1, \widehat{u}_1) \leq \frac{1}{\lambda_1 - \lambda_2} \langle \Sigma - S, u_1 u_1^T - \widehat{u}_1 \widehat{u}_1^T \rangle = \frac{1}{\lambda_1 - \lambda_2} \langle \Sigma - S, P(u_1 u_1^T - \widehat{u}_1 \widehat{u}_1^T) P \rangle$$
$$= \frac{1}{\lambda_1 - \lambda_2} \langle P(\Sigma - S)P, u_1 u_1^T - \widehat{u}_1 \widehat{u}_1^T \rangle \leq \frac{1}{\lambda_1 - \lambda_2} \|P(\Sigma - S)P\|_2 \|u_1 u_1^T - \widehat{u}_1 \widehat{u}_1^T\|_*$$
$$= \frac{2}{\lambda_1 - \lambda_2} |\sin \angle (u_1, \widehat{u}_1)| \|P(\Sigma - S)P\|_2.$$

Here we remind that $\|\cdot\|_*$ is defined as the sum of singular values. The second inequality follows from Lemma A.1.1 of Vu and Lei (2012a), which states that $u_1 u_1^T - \widehat{u}_1 \widehat{u}_1^T$ has the following singular values

$$|\sin \angle (u_1, \widehat{u}_1)|, |\sin \angle (u_1, \widehat{u}_1)|, 0, \ldots, 0.$$

The conclusion follows from

$$\|P(\Sigma - S)P\|_2 \leq \sup_{v \in \mathbb{S}^{d-1} \cap \mathbb{B}_0(2s)} |v^T (\Sigma - S) v|.$$

$\square$

The following lemma from Vershynin (2010) is about the $\epsilon$-net arguments and will be useful later.

**Lemma .2.** An $\epsilon$-net $\mathbb{N}_\epsilon$ of a sphere $\mathbb{S}^{n-1}$ (equipped with Euclidean distance) is a subset of $\mathbb{S}^{n-1}$, if for any $v \in \mathbb{S}^{n-1}$, there is $u \in \mathbb{N}_\epsilon$, subject to $\|u-v\|_2 \leq \epsilon$. For any $\epsilon > 0$, $|\mathbb{N}_\epsilon| \leq (1+2/\epsilon)^n$. Moreover, for any matrix $B \in \mathbb{R}^{n \times n}$, the following inequality holds for $\epsilon \in (0, 1/2)$

$$\sup_{v_1 \in \mathbb{S}^{n-1}} |v_1^T B v_1| \leq (1-2\epsilon)^{-1} \sup_{v_2 \in \mathbb{N}_\epsilon} |v_2^T B v_2|.$$

The next lemma comes from Ledoux and Talagrand (2011) and is useful for the proof of the main theorem.

**Lemma .3.** Let $x_1, \ldots, x_T \in \mathbb{R}$ be independently and identically drawn from distribution $N(0,1)$ and $X = (x_1, \ldots, x_T)^T \in \mathbb{R}^T$ be a random vector. A mapping $f$ from $\mathbb{R}^T$ to $\mathbb{R}$ is Lipschitz, i.e., for any $v_1, v_2 \in \mathbb{R}^T$, there exists $L \geq 0$ such that $|f(v_1) - f(v_2)| \leq L\|v_1 - v_2\|_2$. Then we have, for any $t > 0$,

$$\mathbb{P}\big\{|f(X) - E\{f(X)\}| > t\big\} \leq 2\exp\left(-\frac{t^2}{2L^2}\right). \tag{.2}$$

With Lemma .2 and Lemma .3, we are able to prove the Theorem 4.1.

*of Theorem 4.1.* First we adapt the proof technique in Lemma 8 of Negahban and Wainwright (2011) to construct a Gaussian random vector $Y$ and a Lipschitz mapping $f$ to invoke Lemma .3. Let $x_1, \ldots, x_T$ be the realizations and $Y = (x_1^T v, \ldots, x_T^T v)^T \in \mathbb{R}^T$ be a zero-mean Gaussian random vector. We denote the covariance matrix of $Y$ as $\Sigma_Y$, which can be decomposed as $\Sigma_Y = Q^T Q$. Here $Q \in \mathbb{R}^{T \times T}$ is a matrix with orthonormal columns. We define $\sigma = \|Q\|_2 = \|\Sigma_Y\|_2^{1/2}$. Then we have

$$v^T S v = \frac{1}{T}\sum_{i=1}^{T} v^T x_i x_i^T v = Y^T Y / T = \big(\|Y\|_2 / T^{1/2}\big)^2,$$

$$v^T \Sigma v = v^T E(XX^T) v = E(Y^T Y / T) = E\left\{\big(\|Y\|_2/T^{1/2}\big)^2\right\}.$$

We consider a zero-mean Gaussian random vector $V = Q^{-1} Y \in \mathbb{R}^T$ which has covariance matrix $I_T$. We define $W = \|Y\|_2 / T^{1/2}$ and $f(v) = \|Qv\|_2 / T^{1/2}, v \in \mathbb{R}^T$. Therefore we have $W = f(V)$. It can be verified that mapping $f$ from $\mathbb{R}^T$ to $\mathbb{R}$ is Lipschitz with $L = \|Q\|_2 / T^{1/2} = \sigma/T^{1/2}$. Using Lemma .3, we know $W$ satisfies

$$\mathbb{P}\big\{|W - E(W)| \geq t\big\} \leq 2\exp\left(-\frac{t^2 T}{2\sigma^2}\right). \tag{.3}$$

Since $E(W^2) \geq 0$, $E(W) \geq 0$ and $\text{Var}(W) = E(W^2) - \{E(W)\}^2 \geq 0$, we have

$$0 \leq \{E(W^2)\}^{1/2} - E(W) \leq \{E(W^2)\}^{1/2} + E(W),$$

which implies

$$0 \leq \left[\{E(W^2)\}^{1/2} - E(W)\right]^2 \leq \left[\{E(W^2)\}^{1/2} + E(W)\right]\left[\{E(W^2)\}^{1/2} - E(W)\right] = \text{Var}(W).$$

From (.3) we have

$$\mathrm{Var}(W) = E\left\{|W-E(W)|^2\right\} = \int_0^\infty \mathbb{P}\left\{|W-E(W)|^2 \geq t^2\right\} d(t^2) \leq \int_0^\infty 2\exp\left(-\frac{t^2 T}{2\sigma^2}\right) d(t^2) = 4\sigma^2/T,$$

which indicates

$$0 \leq \{E(W^2)\}^{1/2} - E(W) \leq \{\mathrm{Var}(W)\}^{1/2} = 2\sigma/T^{1/2}. \tag{.4}$$

Since $|W - E(W)| \leq t$ together with (.4) implies $|W - \{E(W^2)\}^{1/2}| \leq t + 2\sigma/T^{1/2}$, we have

$$\begin{aligned}
\mathbb{P}\left\{\left|(v^{\mathrm{T}}\Sigma v)^{1/2} - (v^{\mathrm{T}} S v)^{1/2}\right| \geq t + 2\sigma/T^{1/2}\right\} &\tag{.5} \\
= \mathbb{P}\left\{\left|W - \{E(W^2)\}^{1/2}\right| \geq t + 2\sigma/T^{1/2}\right\} & \\
\leq \mathbb{P}\left\{|W - E(W)| \geq t\right\} & \\
\leq 2\exp\left(-\frac{t^2 T}{2\sigma^2}\right). &
\end{aligned}$$

Also we have $(v^{\mathrm{T}}\Sigma v)^{1/2} = \{E(W^2)\}^{1/2} = \{\mathrm{tr}(QQ^{\mathrm{T}})/T\}^{1/2} = \|Q\|_F/T^{1/2} \leq \sigma$. Then using (.5) we can get

$$\begin{aligned}
\mathbb{P}\left\{\left|(v^{\mathrm{T}}\Sigma v)^{1/2} + (v^{\mathrm{T}} S v)^{1/2}\right| \geq t + 4\sigma\right\} & \\
\leq \mathbb{P}\left\{\left|(v^{\mathrm{T}} S v)^{1/2} + (v^{\mathrm{T}}\Sigma v)^{1/2}\right| \geq t + 2\sigma/T^{1/2} + 2(v^{\mathrm{T}}\Sigma v)^{1/2}\right\} & \\
\leq \mathbb{P}\left\{\left|(v^{\mathrm{T}}\Sigma v)^{1/2} - (v^{\mathrm{T}} S v)^{1/2}\right| \geq t + 2\sigma/T^{1/2}\right\} & \\
\leq 2\exp\left(-\frac{t^2 T}{2\sigma^2}\right). &
\end{aligned}$$

For any $t_1, t_2 > 0$, we have

$$\mathbb{P}\left\{\left|v^{\mathrm{T}}(\Sigma - S)v\right| \geq (t_1 + 2\sigma/T^{1/2})(t_2 + 4\sigma)\right\} \leq 2\exp\left(-\frac{t_1^2 T}{2\sigma^2}\right) + 2\exp\left(-\frac{t_2^2 T}{2\sigma^2}\right).$$

We assume there is a fixed subset $\mathbb{K} \subset \{1, \ldots, d\}$ with cardinality $2s$ and define $\mathbb{B}_{\mathbb{K}} = \{v | \text{for any } i \in \{1, \ldots, d\} \setminus \mathbb{K}, v_i = 0\}$. For any $t_1, t_2 > 0$, we define event $\mathcal{E}_{\mathbb{K}}$, $\mathcal{E}_v$ and $\mathcal{E}$ as

$$\begin{aligned}
\mathcal{E}_{\mathbb{K}} &= \left\{\sup_{v \in \mathbb{S}^{d-1} \cap \mathbb{B}_{\mathbb{K}}} |v^{\mathrm{T}}(\Sigma - S)v| \geq 2(t_1 + 2\sigma/T^{1/2})(t_2 + 4\sigma)\right\}, \\
\mathcal{E}_v &= \left\{|v^{\mathrm{T}}(\Sigma - S)v| \geq (t_1 + 2\sigma/T^{1/2})(t_2 + 4\sigma)\right\}, \\
\mathcal{E} &= \left\{\sup_{v \in \mathbb{S}^{d-1} \cap \mathbb{B}_0(2s)} |v^{\mathrm{T}}(\Sigma - S)v| \geq 2(t_1 + 2\sigma/T^{1/2})(t_2 + 4\sigma)\right\}.
\end{aligned}$$

According to Lemma .2, we define the $1/4$-net of $\mathbb{S}^{d-1} \cap \mathbb{B}_{\mathbb{K}}$ as $\mathbb{N}_{\mathbb{K}}$ and obtain, for $|\mathbb{K}| = 2s$

$$\mathbb{P}(\mathcal{E}) \leq \binom{d}{2s}\mathbb{P}(\mathcal{E}_{\mathbb{K}}) \leq \binom{d}{2s}|\mathbb{N}_{\mathbb{K}}|\mathbb{P}(\mathcal{E}_v) \leq \binom{d}{2s} 9^{2s}\left\{2\exp\left(-\frac{t_1^2 T}{2\sigma^2}\right) + 2\exp\left(-\frac{t_2^2 T}{2\sigma^2}\right)\right\}.$$

Now we derive the upper bound for $\sigma^2$ with the proof technique adapted from Lemma 17 of Loh and Wainwright (2012),

$$\sigma^2 = \|\Sigma_Y\|_2 \leq \|\Sigma_Y\|_\infty = \max_{i \in \{1,\ldots,T\}} \sum_{j=1}^T |(\Sigma_Y)_{ij}|.$$

Since $(\Sigma_Y)_{ij} = v^{\mathrm{T}} \operatorname{Cov}(x_i, x_j) v = v^{\mathrm{T}} \operatorname{Cov}(x_j, x_i) v$, with (2.4) we have

$$\sum_{j=1}^T |(\Sigma_Y)_{ij}| = \sum_{j \neq i} |v^{\mathrm{T}} \operatorname{Cov}(x_i, x_j) v| + |v^{\mathrm{T}} \operatorname{Cov}(x_i, x_i) v| = \sum_{j \neq i} |v^{\mathrm{T}} A^{|i-j|} \Sigma v| + |v^{\mathrm{T}} \Sigma v|$$

$$\leq \sum_{j \neq i} \|\Sigma\|_2 \|A\|_2^{|i-j|} + \|\Sigma\|_2 \leq 2\|\Sigma\|_2/(1 - \|A\|_2),$$

where $\|A\|_2 < 1$. We define $\delta^2 = 2\|\Sigma\|_2/(1 - \|A\|_2) \geq \sigma^2$. Then for any $t_1, t_2 > 0$, we have

$$\mathbb{P}\left\{\sup_{v \in \mathbb{S}^{d-1} \cap \mathbb{B}_0(2s)} |v^{\mathrm{T}}(\Sigma - S)v| \geq 2(t_1 + 2\delta/T^{1/2})(t_2 + 4\delta)\right\} \leq \binom{d}{2s} 9^{2s} \left\{2\exp\left(-\frac{t_1^2 T}{2\delta^2}\right) + 2\exp\left(-\frac{t_2^2 T}{2\delta^2}\right)\right\}.$$

Letting $t_1 = \{s\delta^2 \log(d)/T\}^{1/2}$ and $t_2$ be a constant, we get

$$\sup_{v \in \mathbb{S}^{d-1} \cap \mathbb{B}_0(2s)} |v^{\mathrm{T}}(\Sigma - S)v| = O_P\left[\left\{\frac{s \log(d)}{T}\left(\frac{\|\Sigma\|_2}{1 - \|A\|_2}\right)\right\}^{1/2}\right]. \tag{.6}$$

Combining Lemma .1 and (.6), we obtain the rate of convergence for estimating the leading eigenvector $u_1$

$$|\sin \angle (u_1, \widehat{u}_1)| = O_P\left[\frac{1}{\lambda_1(\Sigma) - \lambda_2(\Sigma)}\left\{\frac{s \log(d)}{T}\left(\frac{\|\Sigma\|_2}{1 - \|A\|_2}\right)\right\}^{1/2}\right],$$

which completes the proof of Theorem 4.1. □

### Proof of Theorem 4.2

We extend the theoretical result in (4.1) to subspace estimation ($m \geq 1$) as in (4.2). Motivated by Lemma .1 in which $\sin \angle (u_1, \widehat{u}_1)$ is quantified, we need to further quantify the distance between the two subspaces $\mathcal{U}$ and $\widehat{\mathcal{U}}$ with $\sin(U, \widehat{U})$ defined in (2.1). In the sequel we define $\Pi = UU^{\mathrm{T}}$ and $\widehat{\Pi} = \widehat{U}\widehat{U}^{\mathrm{T}}$ to be the projection matrices of subspaces $\mathcal{U}$ and $\widehat{\mathcal{U}}$ respectively. The following two lemmas come from Vu and Lei (2012b) and provide useful properties of $\sin(U, \widehat{U})$. We provide the proofs for completeness.

**Lemma .4.** For $U$ and $\widehat{U}$, we have $\|\sin(U, \widehat{U})\|_F = \|\Pi \widehat{\Pi}^\perp\|_F = \|\Pi^\perp \widehat{\Pi}\|_F$.

*Proof.* According to the definition in (2.1), we have

$$\sin(U, \widehat{U}) = \sin\{\Theta(\mathcal{U}, \widehat{\mathcal{U}})\} = \text{diag}\{\sin(\theta_1), \ldots, \sin(\theta_m)\},$$

where $\theta_k = \arcsin\{\sigma_k(\Pi\widehat{\Pi}^\perp)\}$. Hence we get $\|\sin(U, \widehat{U})\|_F = \|\Pi\widehat{\Pi}^\perp\|_F$. According to symmetry, we know

$$\|\sin(U, \widehat{U})\|_F = \|\sin(\widehat{U}, U)\|_F = \|\widehat{\Pi}\Pi^\perp\|_F = \|\Pi^\perp\widehat{\Pi}\|_F,$$

which completes the proof. □

The following lemma bounds $\|\sin(U, \widehat{U})\|_F^2$ with $\langle S - \Sigma, \widehat{U}\widehat{U}^T - UU^T \rangle$ and forms the basis for our proofs.

**Lemma .5.** *$U$ and $\widehat{U}$ satisfy*

$$\|\sin(U, \widehat{U})\|_F^2 \leq \frac{\langle S - \Sigma, \widehat{U}\widehat{U}^T - UU^T \rangle}{\lambda_m(\Sigma) - \lambda_{m+1}(\Sigma)}. \tag{.7}$$

*Proof.* We utilize the fact $\Pi = \sum_{i=1}^m u_i u_i^T$ and $\Pi^\perp = \sum_{i=m+1}^d u_i u_i^T$, where $u_i$ is the $i$-th eigenvector of $\Sigma$, $\Sigma u_i = \lambda_i(\Sigma) u_i$ and $u_i^T u_i = 1$. We have

$$\langle \Sigma, UU^T - \widehat{U}\widehat{U}^T \rangle = \langle \Sigma, \Pi - \widehat{\Pi} \rangle = \langle \Sigma, \Pi(I_d - \widehat{\Pi}) - (I_d - \Pi)\widehat{\Pi} \rangle$$

$$= \langle \Pi\Sigma, \widehat{\Pi}^\perp \rangle - \langle \Pi^\perp\Sigma, \widehat{\Pi} \rangle = \sum_{i=1}^m \lambda_i(\Sigma)\langle u_i u_i^T, \widehat{\Pi}^\perp \rangle - \sum_{i=m+1}^d \lambda_i(\Sigma)\langle u_i u_i^T, \widehat{\Pi} \rangle.$$

Since $\lambda_i(\Sigma)$ is the $i$-th largest eigenvalue of $\Sigma$ and $\langle u_i u_i^T, \widehat{\Pi}^\perp \rangle \geq 0, \langle u_i u_i^T, \widehat{\Pi} \rangle \geq 0$, we have

$$\sum_{i=1}^m \lambda_i(\Sigma)\langle u_i u_i^T, \widehat{\Pi}^\perp \rangle - \sum_{i=m+1}^d \lambda_i(\Sigma)\langle u_i u_i^T, \widehat{\Pi} \rangle$$

$$\geq \lambda_m(\Sigma) \sum_{i=1}^m \langle u_i u_i^T, \widehat{\Pi}^\perp \rangle - \lambda_{m+1}(\Sigma) \sum_{i=m+1}^d \langle u_i u_i^T, \widehat{\Pi} \rangle = \lambda_m(\Sigma)\langle \Pi, \widehat{\Pi}^\perp \rangle - \lambda_{m+1}(\Sigma)\langle \Pi^\perp, \widehat{\Pi} \rangle.$$

Due to the idempotence of the projection operators, we have

$$\lambda_m(\Sigma)\langle \Pi, \widehat{\Pi}^\perp \rangle - \lambda_{m+1}(\Sigma)\langle \Pi^\perp, \widehat{\Pi} \rangle = \lambda_m(\Sigma)\langle \Pi\widehat{\Pi}^\perp, \Pi\widehat{\Pi}^\perp \rangle - \lambda_{m+1}(\Sigma)\langle \Pi^\perp\widehat{\Pi}, \Pi^\perp\widehat{\Pi} \rangle$$

$$= \{\lambda_m(\Sigma) - \lambda_{m+1}(\Sigma)\}\|\Pi\widehat{\Pi}^\perp\|_F^2 = \{\lambda_m(\Sigma) - \lambda_{m+1}(\Sigma)\}\|\sin(U, \widehat{U})\|_F^2,$$

where the second equality comes from $\|\widehat{\Pi}\Pi^\perp\|_F = \|\Pi^\perp\widehat{\Pi}\|_F$ in Lemma .4.

Hence we obtain

$$\langle \Sigma, UU^T - \widehat{U}\widehat{U}^T \rangle \geq \{\lambda_m(\Sigma) - \lambda_{m+1}(\Sigma)\}\|\sin(U, \widehat{U})\|_F^2. \tag{.8}$$

Since $\widehat{U}$ is the minimizer of (3.3), we know $\langle S, \widehat{U}\widehat{U}^T \rangle \geq \langle S, UU^T \rangle$. Adding $\langle S, \widehat{U}\widehat{U}^T - UU^T \rangle \geq 0$ to the both sides of (.8), we have

$$\langle S - \Sigma, \widehat{U}\widehat{U}^T - UU^T \rangle \geq \{\lambda_m(\Sigma) - \lambda_{m+1}(\Sigma)\}\|\sin(U, \widehat{U})\|_F^2.$$

Since we assume $\lambda_m(\Sigma) \neq \lambda_{m+1}(\Sigma)$, we complete the proof. □

The following lemma further bounds $\|\sin(U,\widehat{U})\|_F^2$.

**Lemma .6.** $U$ and $\widehat{U}$ satisfy

$$\|\sin(U,\widehat{U})\|_F^2 \le \frac{\|\widehat{U}\widehat{U}^{\mathrm{T}} - UU^{\mathrm{T}}\|_F \sup_{v\in\mathbb{S}^{d-1}\cap\mathbb{B}_0(2ms)} |v^{\mathrm{T}}(S-\Sigma)v|}{\lambda_m(\Sigma) - \lambda_{m+1}(\Sigma)},$$

where $\mathbb{S}^{d-1}$ is the $d$-dimensional Euclidean unit sphere and $\mathbb{B}_0(2ms)$ is the $\ell_0$ ball with radius $2ms$.

*Proof.* Since we assume $\mathcal{U},\widehat{\mathcal{U}} \in \mathcal{P}_0(s)$, i.e., $\|U\|_0 \le s, \|\widehat{U}\|_0 \le s$, according to the definition of $\|\cdot\|_0$ for a matrix in (3.1), there are at least $d-s$ all-zero rows in both $U$ and $\widehat{U}$. In other words, there are at most $ms$ nonzero entries, i.e., $\sum_{i=1}^m \|u_i\|_0 \le ms$, $\sum_{i=1}^m \|\widehat{u}_i\|_0 \le ms$. Now we define a diagonal matrix $P = \mathrm{diag}\{p_1,\ldots,p_d\}$. Let $p_i = 1$ if and only if $U_{i,*} \ne 0$ or $\widehat{U}_{i,*} \ne 0$, where $U_{i,*}$ and $\widehat{U}_{i,*}$ are the $i$-th rows of $U$ and $\widehat{U}$ respectively. Otherwise, let $p_i = 0$. Thus we have $PU = U$, $P\widehat{U} = \widehat{U}$. There are at least $2ms$ nonzero entries in the diagonal of $P$. Using Lemma .5, we have

$$\|\sin(U,\widehat{U})\|_F^2 \le \frac{\langle S - \Sigma, \widehat{U}\widehat{U}^{\mathrm{T}} - UU^{\mathrm{T}}\rangle}{\lambda_m(\Sigma) - \lambda_{m+1}(\Sigma)} = \frac{\langle S - \Sigma, P(\widehat{U}\widehat{U}^{\mathrm{T}} - UU^{\mathrm{T}})P\rangle}{\lambda_m(\Sigma) - \lambda_{m+1}(\Sigma)}.$$

Since $\|\cdot\|_2$ and $\|\cdot\|_*$ are dual norms, we get

$$\|\sin(U,\widehat{U})\|_F^2 = \frac{\langle P(S-\Sigma)P, \widehat{U}\widehat{U}^{\mathrm{T}} - UU^{\mathrm{T}}\rangle}{\lambda_m(\Sigma) - \lambda_{m+1}(\Sigma)} \le \frac{\|P(S-\Sigma)P\|_2 \|UU^{\mathrm{T}} - \widehat{U}\widehat{U}^{\mathrm{T}}\|_*}{\lambda_m(\Sigma) - \lambda_{m+1}(\Sigma)}.$$

According to the definition of $P$ and $\|\cdot\|_2$, we have

$$\|P(S-\Sigma)P\|_2 \le \sup_{v\in\mathbb{S}^{d-1}\cap\mathbb{B}_0(2ms)} |v^{\mathrm{T}}(S-\Sigma)v|,$$

which completes the proof. □

Since we can bound $\sup_{v\in\mathbb{S}^{d-1}\cap\mathbb{B}_0(2ms)} |v^{\mathrm{T}}(S-\Sigma)v|$ with the similar argument as in the proof of Theorem 4.1, we only need to bound $\|UU^{\mathrm{T}} - \widehat{U}\widehat{U}^{\mathrm{T}}\|_*$ in the following lemma.

**Lemma .7.** We remind that $\|\cdot\|_*$ is defined as the sum of singular values. For $U$ and $\widehat{U}$, we have

$$\|UU^{\mathrm{T}} - \widehat{U}\widehat{U}^{\mathrm{T}}\|_* \le 2m^{1/2}\|\sin(U,\widehat{U})\|_F.$$

*Proof.* We construct the auxiliary semi-orthonormal matrices $U' \in \mathbb{R}^{d\times(d-m)}$ and $\widehat{U}' \in \mathbb{R}^{d\times(d-m)}$ so that $(U,\ U')$ and $(\widehat{U},\ \widehat{U}')$ are orthonormal matrices. We define

$$G = \begin{pmatrix} U & U' \end{pmatrix}^{\mathrm{T}} \begin{pmatrix} \widehat{U} & \widehat{U}' \end{pmatrix} \in \mathbb{R}^{d\times d},$$

which is also an orthonormal matrix.

According to the procedure of Cosine-Sine decomposition (Stewart and Sun, 1990; Paige and Wei, 1994), there exist orthonormal matrices $M$ and $N$ such that

$$M^{\mathrm{T}} G N = \begin{pmatrix} \widetilde{\Gamma} & -\Gamma & 0 \\ \Gamma & \widetilde{\Gamma} & 0 \\ 0 & 0 & I_{d-2m} \end{pmatrix}, \tag{.9}$$

where

$$M = \begin{pmatrix} M_1 & 0 \\ 0 & M_2 \end{pmatrix}, \quad N = \begin{pmatrix} N_1 & 0 \\ 0 & N_2 \end{pmatrix},$$

and $M_1, N_1 \in \mathbb{R}^{m \times m}, M_2, N_2 \in \mathbb{R}^{(d-m) \times (d-m)}$ are all orthonormal matrices. Cosine-Sine decomposition guarantees that $\widetilde{\Gamma} = \mathrm{diag}(\widetilde{\gamma}_1, \ldots, \widetilde{\gamma}_m) \in \mathbb{R}^{m \times m}$ with $\widetilde{\gamma}_1 \geq \cdots \geq \widetilde{\gamma}_m > 0$ and $\Gamma = \mathrm{diag}(\gamma_1, \ldots, \gamma_m) \in \mathbb{R}^{m \times m}$ with $\gamma_1 \geq \cdots \geq \gamma_m > 0$. Moreover, we have

$$\widetilde{\Gamma}^2 + \Gamma^2 = I_m. \tag{.10}$$

Apparently (.9) is equivalent to

$$\begin{pmatrix} UM_1 & U'M_2 \end{pmatrix}^{\mathrm{T}} \begin{pmatrix} \widehat{U}N_1 & \widehat{U}'N_2 \end{pmatrix} = \begin{pmatrix} \widetilde{\Gamma} & -\Gamma & 0 \\ \Gamma & \widetilde{\Gamma} & 0 \\ 0 & 0 & I_{d-2m} \end{pmatrix}.$$

By comparing each entry in the left and right hand sides of the equality above, we obtain

$$HUM_1 = \begin{pmatrix} I_m \\ 0 \end{pmatrix} \in \mathbb{R}^{d \times m}, \quad H\widehat{U}N_1 = \begin{pmatrix} \widetilde{\Gamma} \\ \Gamma \\ 0 \end{pmatrix} \in \mathbb{R}^{d \times m},$$

where $\Gamma, \widetilde{\Gamma} \in \mathbb{R}^{m \times m}$. Here $H = \begin{pmatrix} UM_1, & U'M_2 \end{pmatrix}^{\mathrm{T}} \in \mathbb{R}^{d \times d}$ is also an orthonormal matrix. Now we consider the singular values of $\Pi \widehat{\Pi}^{\perp} = UU^{\mathrm{T}}(I_d - \widehat{U}\widehat{U}^{\mathrm{T}})$. With (.10) we obtain

$$\begin{aligned}
H(UU^{\mathrm{T}})(I_d - \widehat{U}\widehat{U}^{\mathrm{T}})H^{\mathrm{T}} &= HUU^{\mathrm{T}}H^{\mathrm{T}} - H(UU^{\mathrm{T}})(\widehat{U}\widehat{U}^{\mathrm{T}})H^{\mathrm{T}} \\
&= HUU^{\mathrm{T}}H^{\mathrm{T}} - (HUU^{\mathrm{T}}H^{\mathrm{T}})(H\widehat{U}\widehat{U}^{\mathrm{T}}H^{\mathrm{T}}) \\
&= (HUM_1)M_1^{\mathrm{T}}U^{\mathrm{T}}H^{\mathrm{T}} - \left\{(HUM_1)M_1^{\mathrm{T}}U^{\mathrm{T}}H^{\mathrm{T}}\right\}\left\{(H\widehat{U}N_1)N_1^{\mathrm{T}}\widehat{U}^{\mathrm{T}}H^{\mathrm{T}}\right\} \\
&= \begin{pmatrix} I_m - \widetilde{\Gamma}\widetilde{\Gamma} & -\widetilde{\Gamma}\Gamma & 0 \\ 0 & 0 & 0 \end{pmatrix} = \begin{pmatrix} \Gamma\Gamma & -\widetilde{\Gamma}\Gamma & 0 \\ 0 & 0 & 0 \end{pmatrix} = \begin{pmatrix} \Gamma \\ 0 \end{pmatrix}\begin{pmatrix} \Gamma & -\widetilde{\Gamma} & 0 \end{pmatrix} \in \mathbb{R}^{d \times d}.
\end{aligned}$$

Since $\begin{pmatrix} \Gamma, & -\widetilde{\Gamma}, & 0 \end{pmatrix} \in \mathbb{R}^{m \times d}$ has orthonormal rows according to (.10), the singular values of $H(UU^{\mathrm{T}})(I_d - \widehat{U}\widehat{U}^{\mathrm{T}})H^{\mathrm{T}}$ are $\gamma_1, \ldots, \gamma_m, 0, \ldots, 0$. Since $H$ is an orthonormal matrix, $UU^{\mathrm{T}}(I_d - \widehat{U}\widehat{U}^{\mathrm{T}}) = \Pi\widehat{\Pi}^{\perp}$ has the same set of singular values.

We turn to consider the singular values of $UU^{\mathrm{T}} - \widehat{U}\widehat{U}^{\mathrm{T}}$. We have

$$H(UU^{\mathrm{T}} - \widehat{U}\widehat{U}^{\mathrm{T}})H^{\mathrm{T}} = HUU^{\mathrm{T}}H^{\mathrm{T}} - H\widehat{U}\widehat{U}^{\mathrm{T}}H^{\mathrm{T}}$$
$$= (HUM_1)M_1^{\mathrm{T}}U^{\mathrm{T}}H^{\mathrm{T}} - (H\widehat{U}N_1)N_1^{\mathrm{T}}\widehat{U}^{\mathrm{T}}H^{\mathrm{T}} = (HUM_1)M_1^{\mathrm{T}}U^{\mathrm{T}}H^{\mathrm{T}} - (H\widehat{U}N_1)N_1^{\mathrm{T}}\widehat{U}^{\mathrm{T}}H^{\mathrm{T}}$$
$$= \begin{pmatrix} I_m - \widetilde{\Gamma}\widetilde{\Gamma} & -\widetilde{\Gamma}\Gamma & 0 \\ -\Gamma\widetilde{\Gamma} & -\Gamma\Gamma & 0 \\ 0 & 0 & 0 \end{pmatrix} = \begin{pmatrix} \Gamma\Gamma & -\widetilde{\Gamma}\Gamma & 0 \\ -\Gamma\widetilde{\Gamma} & -\Gamma\Gamma & 0 \\ 0 & 0 & 0 \end{pmatrix} \in \mathbb{R}^{d\times d}.$$

As we have mentioned, $\widetilde{\Gamma}, \Gamma \in \mathbb{R}^{m\times m}$ are diagonal matrices. We can easily obtain that the singular values of $H(UU^{\mathrm{T}} - \widehat{U}\widehat{U}^{\mathrm{T}})H^{\mathrm{T}}$ are $\gamma_1, \gamma_1, \ldots, \gamma_m, \gamma_m, 0, \ldots, 0$. Since $H$ is an orthonormal matrix, $UU^{\mathrm{T}} - \widehat{U}\widehat{U}^{\mathrm{T}}$ has the same set of singular values. We obtain that

$$\|UU^{\mathrm{T}} - \widehat{U}\widehat{U}^{\mathrm{T}}\|_* = 2\sum_{i=1}^m \gamma_i \le 2m^{1/2}\left(\sum_{i=1}^m \gamma_i^2\right)^{1/2}.$$

By our definitions, we have

$$\gamma_k = \sigma_k(\Pi\widehat{\Pi}^\perp) = \sin\{\theta_k(\mathcal{U},\widehat{\mathcal{U}})\} \quad (k = 1, \ldots, m).$$

Therefore, since $\|\sin(U,\widehat{U})\|_F = \left(\sum_{i=1}^m \gamma_i^2\right)^{1/2}$, we have

$$\|UU^{\mathrm{T}} - \widehat{U}\widehat{U}^{\mathrm{T}}\|_* \le 2m^{1/2}\|\sin(U,\widehat{U})\|_F,$$

which completes the proof. □

Now we are able to prove the rate of convergence for subspace estimation in Theorem 4.2.

*of Theorem 4.2.* From Lemma .6 and Lemma .7 we have

$$\|\sin(U,\widehat{U})\|_F \le 2m^{1/2}\frac{\sup_{v\in\mathbb{S}^{d-1}\cap\mathbb{B}_0(2ms)}|v^{\mathrm{T}}(S-\Sigma)v|}{\lambda_m(\Sigma) - \lambda_{m+1}(\Sigma)}. \tag{.11}$$

Following the similar argument as in the proof of Theorem 4.1, we obtain the rate of convergence of $\sup_{v\in\mathbb{S}^{d-1}\cap\mathbb{B}_0(2ms)}|v^{\mathrm{T}}(S-\Sigma)v|$ by replacing $s$ with $ms$ in (.6). Combining Lemma .7 and (.11), we get

$$\|\sin(U,\widehat{U})\|_F = O_P\left[\frac{m}{\lambda_m(\Sigma) - \lambda_{m+1}(\Sigma)}\left\{\frac{s\log(d)}{T}\left(\frac{\|\Sigma\|_2}{1-\|A\|_2}\right)\right\}^{1/2}\right],$$

which completes the proof of Theorem 4.2. □

**Proof of Corollary 4.4**

*of Corollary 4.4.* We define the following random vector, transition matrix, and noise vector,

$$\widetilde{X}_t = \begin{pmatrix} X_t \\ X_{t-1} \\ \vdots \\ X_{t-p+1} \end{pmatrix} \in \mathbb{R}^{pd}, \widetilde{A} = \begin{pmatrix} A_1 & A_2 & \cdots & A_{p-1} & A_p \\ I_d & 0 & \cdots & 0 & 0 \\ 0 & I_d & \cdots & 0 & 0 \\ \vdots & \ddots & \ddots & \ddots & \vdots \\ 0 & 0 & \cdots & I_d & 0 \end{pmatrix} \in \mathbb{R}^{pd \times pd}, \widetilde{Z}_t = \begin{pmatrix} Z_t \\ 0 \\ \vdots \\ 0 \end{pmatrix} \in \mathbb{R}^{pd}, \quad (.12)$$

the lag $p$ vector autoregressive model in (2.5) can be reformulated as the lag 1 vector autoregressive model,

$$\widetilde{X}_{t+1} = \widetilde{A}\widetilde{X}_t + \widetilde{Z}_{t+1} \quad (t \in \mathbb{Z}). \quad (.13)$$

It is easy to show that the sequence $\{\widetilde{X}_t\}_{-\infty}^{\infty}$ in (.12) is stationary. We define the marginal covariance matrix of $(\widetilde{X}_t)_{t=-\infty}^{\infty}$ as $\widetilde{\Sigma}$:

$$\widetilde{\Sigma} = \begin{pmatrix} \Sigma & \text{Cov}(X_1, X_2) & \cdots & \text{Cov}(X_1, X_p) \\ \text{Cov}(X_2, X_1) & \Sigma & \cdots & \vdots \\ \vdots & \ddots & \ddots & \vdots \\ \text{Cov}(X_p, X_1) & \cdots & \cdots & \Sigma \end{pmatrix} \in \mathbb{R}^{pd \times pd}, \quad (.14)$$

and the marginal covariance matrix of $(\widetilde{Z}_t)_{t=-\infty}^{\infty}$ as

$$\widetilde{\Psi} = \begin{pmatrix} \Psi & \cdots & 0 \\ \vdots & \ddots & \vdots \\ 0 & \cdots & 0 \end{pmatrix} \in \mathbb{R}^{pd \times pd}. \quad (.15)$$

The sample covariance matrix $\widetilde{S}$ of $\widetilde{X}_1, \ldots, \widetilde{X}_T$ from $(\widetilde{X}_t)_{t=-\infty}^{\infty}$ is

$$\widetilde{S} = \begin{pmatrix} S & \frac{1}{T}\sum_{i=1}^{T} x_i x_{i+1}^{\text{T}} & \cdots & \frac{1}{T}\sum_{i=1}^{T} x_i x_{i+p-1}^{\text{T}} \\ \frac{1}{T}\sum_{i=1}^{T} x_{i+1} x_i^{\text{T}} & \frac{1}{T}\sum_{i=1}^{T} x_{i+1} x_{i+1}^{\text{T}} & \cdots & \vdots \\ \vdots & \ddots & \ddots & \vdots \\ \frac{1}{T}\sum_{i=1}^{T} x_{i+p-1} x_i & \cdots & \cdots & \frac{1}{T}\sum_{i=1}^{T} x_{i+p-1} x_{i+p-1}^{\text{T}} \end{pmatrix} \in \mathbb{R}^{pd \times pd},$$

where $X_{T+1}, \ldots, X_{T+p-1}$ are also from the underlying vector autoregressive process $(\widetilde{X}_t)_{t=-\infty}^{\infty}$ and $x_{T+1}, \ldots, x_{T+p-1}$ are the realizations of $X_{T+1}, \ldots, X_{T+p-1}$. It is important to remind that we only use $x_{T+1}, \ldots, x_{T+p-1}$ in our proof, not in the estimation procedure.

It is easy to verify that we still have Lemma .6 and Lemma .7. Therefore we get

$$\left\|\sin\left(U,\widehat{U}\right)\right\|_F \leq 2m^{1/2}\frac{\sup_{v\in\mathbb{S}^{d-1}\cap\mathbb{B}_0(2ms)}\left|v^{\mathrm{T}}(S-\Sigma)v\right|}{\lambda_m(\Sigma)-\lambda_{m+1}(\Sigma)} \tag{.16}$$

$$= 2m^{1/2}\frac{\sup_{\widetilde{v}\in\mathbb{S}^{pd-1}\cap\mathbb{B}_0(2ms)\cap\mathbb{G}}\left|\widetilde{v}^{\mathrm{T}}(\widetilde{S}-\widetilde{\Sigma})\widetilde{v}\right|}{\lambda_m(\Sigma)-\lambda_{m+1}(\Sigma)}.$$

$$\leq 2m^{1/2}\frac{\sup_{\widetilde{v}\in\mathbb{S}^{pd-1}\cap\mathbb{B}_0(2ms)}\left|\widetilde{v}^{\mathrm{T}}(\widetilde{S}-\widetilde{\Sigma})\widetilde{v}\right|}{\lambda_m(\Sigma)-\lambda_{m+1}(\Sigma)}.$$

where $\mathbb{G} = \{\widetilde{v} : \widetilde{v}_i = 0,\ i > d\}$ denotes the set of $\widetilde{v} \in \mathbb{R}^{pd}$ with all its $i$-th ($i > d$) entries being zero. Therefore the equality in (.16) holds.

Since $\widetilde{X}_1,\ldots,\widetilde{X}_T$ are from the $pd$-dimensional lag-one autoregressive process $(\widetilde{X}_t)_{t=-\infty}^{\infty}$ with transition matrix $\widetilde{A}$ defined in (.12) and covariance matrix $\widetilde{\Sigma}$ defined in (.14). Following the similar argument as in the proof of Theorem 4.1 and replacing $d$, $A$, $\Sigma$ with $pd$, $\widetilde{A}$ and $\widetilde{\Sigma}$ accordingly in (.6), we have

$$\sup_{\widetilde{v}\in\mathbb{S}^{pd-1}\cap\mathbb{B}_0(2ms)}\left|\widetilde{v}^{\mathrm{T}}(\widetilde{S}-\widetilde{\Sigma})\widetilde{v}\right| = O_P\left[\left\{\frac{ms\log(pd)}{T}\left(\frac{\|\widetilde{\Sigma}\|_2}{1-\|A\|_2}\right)\right\}^{1/2}\right],$$

which together with (.16) implies

$$\left\|\sin\left(U,\widehat{U}\right)\right\|_F = O_P\left[\frac{m}{\lambda_m(\Sigma)-\lambda_{m+1}(\Sigma)}\left\{\frac{s\log(pd)}{T}\left(\frac{\|\widetilde{\Sigma}\|_2}{1-\|\widetilde{A}\|_2}\right)\right\}^{1/2}\right].$$

This completes the proof of Corollary 4.4. □

## Appendix B.

### Experimental Results

In this section, we provide the experimental results in Table 1 and Table 2, where the standard deviations are present in the parentheses. The data are visualized in Figure 1–4 correspondingly.

| $d$ | 64 | 128 | 256 | 64 | 128 | 256 | 64 | 128 | 256 | 64 | 128 | 256 |
|---|---|---|---|---|---|---|---|---|---|---|---|---|
| $\|A\|_2$ | 0 | 0 | 0 | 0.2 | 0.2 | 0.2 | 0.4 | 0.4 | 0.4 | 0.6 | 0.6 | 0.6 |
| $T = 16$ | 0.79 | 0.87 | 0.91 | 0.83 | 0.89 | 0.93 | 0.88 | 0.94 | 0.96 | 0.92 | 0.96 | 0.98 |
| | (0.035) | (0.031) | (0.024) | (0.032) | (0.031) | (0.021) | (0.029) | (0.018) | (0.017) | (0.021) | (0.014) | (0.009) |
| $T = 18$ | 0.78 | 0.85 | 0.91 | 0.81 | 0.89 | 0.92 | 0.86 | 0.93 | 0.96 | 0.92 | 0.96 | 0.98 |
| | (0.037) | (0.035) | (0.026) | (0.037) | (0.028) | (0.024) | (0.030) | (0.021) | (0.018) | (0.019) | (0.015) | (0.008) |
| $T = 20$ | 0.74 | 0.83 | 0.88 | 0.80 | 0.87 | 0.93 | 0.85 | 0.90 | 0.94 | 0.91 | 0.96 | 0.97 |
| | (0.041) | (0.037) | (0.033) | (0.038) | (0.031) | (0.025) | (0.032) | (0.027) | (0.021) | (0.022) | (0.012) | (0.012) |
| $T = 30$ | 0.60 | 0.68 | 0.78 | 0.65 | 0.73 | 0.81 | 0.76 | 0.84 | 0.91 | 0.86 | 0.93 | 0.96 |
| | (0.048) | (0.047) | (0.044) | (0.047) | (0.048) | (0.041) | (0.039) | (0.035) | (0.025) | (0.032) | (0.023) | (0.016) |
| $T = 40$ | 0.47 | 0.56 | 0.63 | 0.53 | 0.61 | 0.70 | 0.66 | 0.75 | 0.83 | 0.81 | 0.89 | 0.94 |
| | (0.048) | (0.053) | (0.056) | (0.047) | (0.051) | (0.052) | (0.046) | (0.045) | (0.041) | (0.038) | (0.026) | (0.022) |
| $T = 50$ | 0.40 | 0.43 | 0.52 | 0.43 | 0.52 | 0.57 | 0.56 | 0.65 | 0.75 | 0.75 | 0.84 | 0.91 |
| | (0.045) | (0.050) | (0.054) | (0.045) | (0.052) | (0.057) | (0.049) | (0.051) | (0.049) | (0.043) | (0.038) | (0.027) |
| $T = 60$ | 0.32 | 0.37 | 0.40 | 0.32 | 0.43 | 0.51 | 0.46 | 0.55 | 0.65 | 0.70 | 0.80 | 0.85 |
| | (0.039) | (0.048) | (0.050) | (0.040) | (0.053) | (0.059) | (0.050) | (0.054) | (0.057) | (0.050) | (0.044) | (0.038) |
| $T = 70$ | 0.26 | 0.28 | 0.34 | 0.27 | 0.35 | 0.39 | 0.41 | 0.47 | 0.59 | 0.64 | 0.72 | 0.82 |
| | (0.039) | (0.041) | (0.050) | (0.034) | (0.046) | (0.054) | (0.047) | (0.053) | (0.056) | (0.049) | (0.049) | (0.043) |
| $T = 80$ | 0.24 | 0.25 | 0.25 | 0.28 | 0.28 | 0.32 | 0.33 | 0.40 | 0.46 | 0.55 | 0.65 | 0.76 |
| | (0.034) | (0.036) | (0.040) | (0.038) | (0.043) | (0.049) | (0.041) | (0.053) | (0.057) | (0.048) | (0.054) | (0.050) |
| $T = 90$ | 0.18 | 0.21 | 0.23 | 0.21 | 0.23 | 0.25 | 0.31 | 0.36 | 0.42 | 0.53 | 0.60 | 0.69 |
| | (0.020) | (0.033) | (0.037) | (0.032) | (0.035) | (0.038) | (0.044) | (0.051) | (0.056) | (0.051) | (0.053) | (0.054) |
| $T = 100$ | 0.17 | 0.18 | 0.19 | 0.19 | 0.20 | 0.24 | 0.24 | 0.28 | 0.35 | 0.44 | 0.54 | 0.67 |
| | (0.020) | (0.022) | (0.030) | (0.025) | (0.032) | (0.040) | (0.034) | (0.038) | (0.054) | (0.050) | (0.054) | (0.057) |
| $T = 150$ | 0.13 | 0.13 | 0.13 | 0.13 | 0.13 | 0.14 | 0.16 | 0.16 | 0.19 | 0.27 | 0.29 | 0.36 |
| | (0.008) | (0.013) | (0.006) | (0.007) | (0.012) | (0.020) | (0.018) | (0.013) | (0.032) | (0.038) | (0.041) | (0.053) |
| $T = 200$ | 0.11 | 0.11 | 0.11 | 0.11 | 0.11 | 0.11 | 0.13 | 0.13 | 0.14 | 0.20 | 0.20 | 0.23 |
| | (0.005) | (0.005) | (0.005) | (0.006) | (0.006) | (0.006) | (0.018) | (0.012) | (0.017) | (0.031) | (0.029) | (0.042) |
| $T = 250$ | 0.10 | 0.10 | 0.10 | 0.10 | 0.10 | 0.10 | 0.11 | 0.11 | 0.11 | 0.16 | 0.15 | 0.16 |
| | (0.005) | (0.005) | (0.005) | (0.005) | (0.005) | (0.005) | (0.005) | (0.006) | (0.007) | (0.020) | (0.017) | (0.025) |
| $T = 300$ | 0.09 | 0.09 | 0.09 | 0.09 | 0.09 | 0.09 | 0.10 | 0.10 | 0.11 | 0.13 | 0.13 | 0.13 |
| | (0.004) | (0.004) | (0.004) | (0.004) | (0.004) | (0.005) | (0.006) | (0.005) | (0.011) | (0.010) | (0.016) | (0.009) |
| $T = 350$ | 0.08 | 0.08 | 0.08 | 0.08 | 0.09 | 0.08 | 0.10 | 0.10 | 0.10 | 0.12 | 0.12 | 0.12 |
| | (0.004) | (0.004) | (0.004) | (0.004) | (0.004) | (0.004) | (0.011) | (0.005) | (0.005) | (0.015) | (0.007) | (0.016) |
| $T = 400$ | 0.08 | 0.08 | 0.08 | 0.08 | 0.08 | 0.08 | 0.09 | 0.09 | 0.09 | 0.11 | 0.11 | 0.11 |
| | (0.004) | (0.004) | (0.003) | (0.004) | (0.004) | (0.004) | (0.004) | (0.004) | (0.004) | (0.006) | (0.006) | (0.007) |
| $T = 450$ | 0.07 | 0.07 | 0.07 | 0.07 | 0.07 | 0.08 | 0.08 | 0.08 | 0.08 | 0.11 | 0.11 | 0.10 |
| | (0.003) | (0.004) | (0.003) | (0.004) | (0.004) | (0.004) | (0.004) | (0.004) | (0.004) | (0.011) | (0.011) | (0.005) |
| $T = 500$ | 0.07 | 0.07 | 0.07 | 0.07 | 0.07 | 0.07 | 0.08 | 0.08 | 0.08 | 0.10 | 0.10 | 0.10 |
| | (0.003) | (0.003) | (0.003) | (0.003) | (0.003) | (0.004) | (0.004) | (0.004) | (0.004) | (0.005) | (0.006) | (0.005) |

Table 1: Empirical error for estimating $u_1$ with $\|A\|_2 = 0, 0.2, \ldots, 0.6$, $d = 64, 128, 256$ and $T = 16, 18, 20, 30, \ldots, 100, 150, \ldots, 500$. The standard deviations are present in the parentheses. The data are visualized in Figure 1 and Figure 3.

| $d$ | 64 | 128 | 256 | 64 | 128 | 256 | 64 | 128 | 256 | 64 | 128 | 256 |
|---|---|---|---|---|---|---|---|---|---|---|---|---|
| $\|A\|_2$ | 0 | 0 | 0 | 0.2 | 0.2 | 0.2 | 0.4 | 0.4 | 0.4 | 0.6 | 0.6 | 0.6 |
| $T=16$ | 1.23 | 1.33 | 1.37 | 1.25 | 1.34 | 1.40 | 1.35 | 1.43 | 1.52 | 1.42 | 1.55 | 1.62 |
| | (0.014) | (0.012) | (0.010) | (0.012) | (0.013) | (0.010) | (0.012) | (0.011) | (0.009) | (0.012) | (0.012) | (0.010) |
| $T=18$ | 1.18 | 1.27 | 1.36 | 1.22 | 1.32 | 1.39 | 1.28 | 1.37 | 1.46 | 1.42 | 1.52 | 1.60 |
| | (0.013) | (0.011) | (0.011) | (0.010) | (0.012) | (0.009) | (0.012) | (0.011) | (0.011) | (0.011) | (0.010) | (0.010) |
| $T=20$ | 1.15 | 1.22 | 1.30 | 1.14 | 1.26 | 1.36 | 1.25 | 1.34 | 1.43 | 1.37 | 1.49 | 1.57 |
| | (0.012) | (0.013) | (0.010) | (0.015) | (0.012) | (0.011) | (0.013) | (0.011) | (0.010) | (0.013) | (0.011) | (0.011) |
| $T=30$ | 0.96 | 1.09 | 1.14 | 1.02 | 1.14 | 1.17 | 1.10 | 1.22 | 1.29 | 1.24 | 1.37 | 1.43 |
| | (0.017) | (0.013) | (0.013) | (0.015) | (0.012) | (0.013) | (0.014) | (0.012) | (0.012) | (0.013) | (0.010) | (0.009) |
| $T=40$ | 0.89 | 0.91 | 1.08 | 0.88 | 0.93 | 1.10 | 1.03 | 1.11 | 1.15 | 1.16 | 1.24 | 1.36 |
| | (0.020) | (0.015) | (0.014) | (0.019) | (0.015) | (0.013) | (0.016) | (0.013) | (0.011) | (0.013) | (0.012) | (0.012) |
| $T=50$ | 0.77 | 0.84 | 0.97 | 0.78 | 0.87 | 1.01 | 0.93 | 1.00 | 1.09 | 1.10 | 1.17 | 1.27 |
| | (0.019) | (0.018) | (0.016) | (0.020) | (0.018) | (0.015) | (0.020) | (0.017) | (0.013) | (0.013) | (0.013) | (0.012) |
| $T=60$ | 0.73 | 0.82 | 0.85 | 0.76 | 0.85 | 0.96 | 0.86 | 0.93 | 1.01 | 1.03 | 1.10 | 1.18 |
| | (0.023) | (0.022) | (0.019) | (0.023) | (0.022) | (0.018) | (0.021) | (0.018) | (0.015) | (0.016) | (0.013) | (0.011) |
| $T=70$ | 0.60 | 0.65 | 0.73 | 0.64 | 0.69 | 0.75 | 0.75 | 0.80 | 0.88 | 0.95 | 1.00 | 1.11 |
| | (0.015) | (0.016) | (0.018) | (0.020) | (0.016) | (0.016) | (0.019) | (0.017) | (0.016) | (0.016) | (0.017) | (0.012) |
| $T=80$ | 0.63 | 0.68 | 0.68 | 0.61 | 0.72 | 0.74 | 0.74 | 0.80 | 0.84 | 0.89 | 0.97 | 1.08 |
| | (0.025) | (0.022) | (0.017) | (0.022) | (0.021) | (0.019) | (0.023) | (0.020) | (0.018) | (0.019) | (0.017) | (0.013) |
| $T=90$ | 0.49 | 0.52 | 0.59 | 0.51 | 0.55 | 0.71 | 0.64 | 0.65 | 0.81 | 0.84 | 0.93 | 0.99 |
| | (0.015) | (0.011) | (0.015) | (0.014) | (0.014) | (0.021) | (0.019) | (0.017) | (0.021) | (0.021) | (0.018) | (0.015) |
| $T=100$ | 0.46 | 0.53 | 0.76 | 0.50 | 0.57 | 0.72 | 0.56 | 0.68 | 0.88 | 0.77 | 0.88 | 1.04 |
| | (0.008) | (0.015) | (0.027) | (0.015) | (0.018) | (0.026) | (0.013) | (0.019) | (0.023) | (0.019) | (0.019) | (0.015) |
| $T=150$ | 0.37 | 0.47 | 0.43 | 0.42 | 0.49 | 0.45 | 0.45 | 0.56 | 0.60 | 0.62 | 0.77 | 0.77 |
| | (0.010) | (0.031) | (0.012) | (0.016) | (0.029) | (0.011) | (0.013) | (0.028) | (0.021) | (0.018) | (0.024) | (0.020) |
| $T=200$ | 0.29 | 0.32 | 0.44 | 0.30 | 0.34 | 0.43 | 0.36 | 0.40 | 0.58 | 0.47 | 0.56 | 0.78 |
| | (0.004) | (0.008) | (0.027) | (0.005) | (0.005) | (0.024) | (0.009) | (0.009) | (0.030) | (0.012) | (0.016) | (0.027) |
| $T=250$ | 0.24 | 0.23 | 0.38 | 0.26 | 0.24 | 0.34 | 0.29 | 0.34 | 0.49 | 0.41 | 0.55 | 0.69 |
| | (0.004) | (0.006) | (0.028) | (0.004) | (0.010) | (0.024) | (0.004) | (0.020) | (0.031) | (0.012) | (0.027) | (0.029) |
| $T=300$ | 0.24 | 0.25 | 0.39 | 0.25 | 0.26 | 0.44 | 0.28 | 0.30 | 0.50 | 0.36 | 0.39 | 0.70 |
| | (0.003) | (0.003) | (0.030) | (0.003) | (0.003) | (0.033) | (0.004) | (0.004) | (0.031) | (0.004) | (0.008) | (0.030) |
| $T=350$ | 0.20 | 0.20 | 0.20 | 0.20 | 0.21 | 0.22 | 0.24 | 0.24 | 0.28 | 0.32 | 0.35 | 0.42 |
| | (0.003) | (0.004) | (0.004) | (0.003) | (0.004) | (0.009) | (0.005) | (0.005) | (0.014) | (0.005) | (0.013) | (0.022) |
| $T=400$ | 0.17 | 0.19 | 0.20 | 0.18 | 0.20 | 0.22 | 0.22 | 0.24 | 0.26 | 0.29 | 0.32 | 0.35 |
| | (0.004) | (0.003) | (0.003) | (0.004) | (0.003) | (0.004) | (0.004) | (0.003) | (0.004) | (0.005) | (0.005) | (0.008) |
| $T=450$ | 0.17 | 0.17 | 0.17 | 0.19 | 0.19 | 0.19 | 0.21 | 0.22 | 0.21 | 0.27 | 0.32 | 0.38 |
| | (0.003) | (0.002) | (0.003) | (0.003) | (0.003) | (0.009) | (0.003) | (0.004) | (0.004) | (0.004) | (0.011) | (0.024) |
| $T=500$ | 0.13 | 0.19 | 0.19 | 0.14 | 0.20 | 0.21 | 0.16 | 0.23 | 0.24 | 0.22 | 0.28 | 0.31 |
| | (0.002) | (0.002) | (0.003) | (0.002) | (0.002) | (0.003) | (0.003) | (0.002) | (0.004) | (0.009) | (0.003) | (0.004) |

Table 2: Empirical error for subspace estimation ($m = 4$) with $\|A\|_2 = 0, 0.2, \ldots, 0.6$, $d = 64, 128, 256$ and $T = 16, 18, 20, 30, \ldots, 100, 150, \ldots, 500$. The standard deviations are present in the parentheses. The data are visualized in Figure 2 and Figure 4.